%% file: REVEL.tex
\documentclass{article}

\usepackage{arxiv}

\usepackage[utf8]{inputenc} % allow utf-8 input
\usepackage[T1]{fontenc}    % use 8-bit T1 fonts
\usepackage{hyperref}       % hyperlinks
\usepackage{url}            % simple URL typesetting
\usepackage{booktabs}       % professional-quality tables
\usepackage{amsfonts}       % blackboard math symbols
\usepackage{nicefrac}       % compact symbols for 1/2, etc.
\usepackage{microtype}      % microtypography
\usepackage{lipsum}		% Can be removed after putting your text content
\usepackage{graphicx}
\usepackage{natbib}
\usepackage{doi}
\usepackage{amsmath}
\usepackage{xcolor}
\usepackage{multirow}
\usepackage{float}
\usepackage{marginnote}
\usepackage{tabularx}

\author{
  Iván Sevillano-García \\
  DaSCI, University of Granada, Granada, 18071, Spain \\
  University of Granada \\
  Granada\\
  isevillano@ugr.es\\
\And
  Julián Luengo-Martín \\
  DaSCI, University of Granada, Granada, 18071, Spain \\
  University of Granada \\
  Granada\\
  jluengo@decsai.ugr.es\\
\And
  Francisco Herrera \\
  Faculty of Computing and Information Technology, King Abdulaziz University, Jeddah, 21589\\
  Saudi Arabia\\
  herrera@decsai.ugr.es
}

\begin{document}

\title{REVEL Framework to measure Local Linear Explanations for black-box models: Deep Learning Image Classification case of study}

%\date{September 9, 1985}	% Here you can change the date presented in the paper title
%\date{} 					% Or removing it

%% Group authors per affiliation:
%\author[1]{Iván Sevillano-García, University of Granada}

%\author[1]{Julián Luengo-Martín, University of Granada}

%\author[1,2]{Francisco Herrera, University of Granada}

%\address[1]{Department of Computer Science and Artificial Intelligence, Andalusian Research Institute in Data Science and Computational Intelligence (DaSCI), University of Granada, Granada, 18071, Spain}

%\address[2]{Faculty of Computing and Information Technology, King Abdulaziz University, Jeddah, 21589, Saudi Arabia}

% Uncomment to remove the date
%\date{}

% Uncomment to override  the `A preprint' in the header
%\renewcommand{\headeright}{Technical Report}
%\renewcommand{\undertitle}{Technical Report}
\renewcommand{\shorttitle}{\textit{arXiv} Template}

%%% Add PDF metadata to help others organize their library
%%% Once the PDF is generated, you can check the metadata with
%%% $ pdfinfo template.pdf

\maketitle

\begin{abstract}
	Explainable artificial intelligence is proposed to provide explanations for reasoning performed by an Artificial Intelligence. There is no consensus on how to evaluate the quality of these explanations, since even the definition of explanation itself is not clear in the literature. In particular, for the widely known Local Linear Explanations, there are qualitative proposals for the evaluation of explanations, although they suffer from theoretical inconsistencies. The case of image is even more problematic, where a visual explanation seems to explain a decision while detecting edges is what it really does. There are a large number of metrics in the literature specialized in quantitatively measuring different qualitative aspects so we should be able to develop metrics capable of measuring in a robust and correct way the desirable aspects of the explanations. In this paper, we propose a procedure called REVEL to evaluate different aspects concerning the quality of explanations with a theoretically coherent development. This procedure has several advances in the state of the art: it standardizes the concepts of explanation and develops a series of metrics not only to be able to compare between them but also to obtain absolute information regarding the explanation itself. The experiments have been carried out on image four datasets as benchmark where we show REVEL's descriptive and analytical power.
\end{abstract}

% keywords can be removed

\input{sections/1_Introduction}

\input{sections/2_RelatedWork}

%\input{sections/3_LEAF_Metrics}

\input{sections/3_Proposal}

\input{sections/4_ExperimentalSetup}

\input{sections/5_Experiments}

\input{sections/6_Conclusion}

\bibliographystyle{unsrtnat}

\section*{Acknowledgements}
This work has been partially supported by the Contract UGR-AM OTRI-6717 and the Contract UGR-AM OTRI-5987.

\bibliography{references}  

\end{document}

%% file: sections/1_Introduction.tex
\section{Introduction}
\label{introduction}

In recent years, Artificial Intelligence (AI) has experienced a huge development, providing solutions to many real-life problems. 
Unfortunately, these systems remain characteristically opaque, which is known as the black-box problem. 
To tackle the comprehension of the black-box, several eXplainable AI (XAI) techniques have been proposed \cite{arrieta2020explainable}. 
In general, the aim is to extract knowledge from black-box models so that they become understandable by a human but it also aims to show the risks of not using the XAI perspective \cite{yan2022towards}.

In the literature, there is a clear separation between model-agnostic and model-specific explanations.
Explanations designed as agnostic do not require knowledge of the model's own structure information \cite{bohanec2017explaining,guidotti2018survey}. 
One of the most used and simple ones are Local Linear Explanation (LLE).% Model-agnostic methods based on LLEs rely on importance attribution, where the most important feature is the feature that influences the classification of the data point the most.}

All proposed explanations are based on different notions of what constitute an explanation and, therefore, are not directly comparable. 
In the literature, there are several proposals to compare explanations. 
In \cite{confalonieri2021historical}, different desirable qualitative aspects for an explanation are proposed, without including ways to measure them.
In \cite{amparore2021trust}, the LEAF framework is proposed, designed for the evaluation and comparison of explanations. 
This framework has 4 different metrics to evaluate different desirable qualitative aspects of explanations. 
However, these metrics have different design inconsistencies which makes them incomplete and biased.

Although there are different measurement proposals, there is no consensus in the XAI literature on how to evaluate explanations since there is no definition of what constitutes a good explanation \cite{gilpin2018explaining}. 
Moreover, these measures have theoretical inconsistencies and, although they are useful to compare explanations, they do not provide absolute information on the explanation itself.
Therefore, a set of robust metrics theoretically correct and representing characteristic behaviors of the method in practice is necessary. 
We also want to emphasize the difficulty of analyzing different factors that must inherently modify the explanation, such as the specific task covered by an AI or the type of data on which the explanation is generated.

Although there is no consensus within the literature on how we should create or even measure explanations, there are different state-of-art tools available that, combined with robust mathematical development, can provide a more generalizable and reliable analysis of the black-box generated explanations.

This work focuses on the proposal of the REVEL framework(Robust Evaluation VEctorized Loca-linear-explanation), whose main contribution is to offer a consistent and theoretically robust analysis of the black-box generated explanations, as well as being useful at a practical level for the evaluation of explanations. 
REVEL takes advantage of the existing state-of-the-art and develops a series of theoretical improvements on the generation and evaluation methods.
In addition, it redefine and propose different quantitative measures to robustly assess different qualitative aspects of the explanations. These measures emerge naturally and are well defined, so that we can extract not only comparative information among explanations but also get an absolute idea about the quality of an explanation on its own.

%This improvement is twofold:}
%\begin{itemize}
    %\item \textcolor{green}{\marginnote{Análisis de conceptos}We conducted an analysis on the state of the art of explanation generation and asked different questions that have not been particularly taken into account, such as the type of data, the normalization of the explanations or the task an artificial intelligence tackles.}
    %\item \textcolor{green}{\marginnote{Desarrollo de buenas medidas}We redefine and propose different quantitative measures to robustly assess different qualitative aspects of the explanations. These measures emerge naturally and are well defined, so that we can extract not only comparative information among explanations but also get an absolute idea about the quality of an explanation on its own.}
%\end{itemize}

Although the theoretical study is generalizable to any kind of data and any kind of task, we focus on image classification in order to simplify the final discussion of the article. In addition, it is easier to work with images for the purpose of the analysis in the article, since it is simpler to generate different number of features with this data type.

The experimental section has been designed to show the analytical and descriptive potential of REVEL. 
%The experiments has been performed on four different widely known image datasets as benchmark. 
%An efficientnet-b2 model has been trained on each of these datasets as a benchmark model from which to extract explanations. 
%After this first training experiment, w
We have designed three different scenarios on which to use REVEL. These scenarios are:

\begin{itemize}
    \item We analyze within LIME how much the number of black box evaluations affects the quality of the explanations.
    % This scenario will provide information on the quality we will be losing by choosing fewer evaluations or the amount of processing time we need to spend generating an explanation in order to get a higher quality explanation.}
    \item Within LIME, we also analyze how the number of features in which we split an image is affecting.
    % This scenario will provide information on the need of increasing or not the granularity of explanation.}
    \item  We compare the two well-known state-of-the-art black box explanation generators, LIME and SHAP, to demonstrate the comparative capability of REVEL.
\end{itemize}

The rest of the paper is organised as follows: Section \ref{background} provides a survey of motivations and basic concepts of LLE and describes two main methods that we will compare, LIME and SHAP. Section \ref{proposal} proposes REVEL framework and highlight its strengths with respect to other methods of evaluating explanations in a theoretical way. Section \ref{experimental_setup} develops a generic experimental pipeline for the comparison of explanations which we use in the following Section \ref{experiments} to perform a comparison of different aspects of LIME and SHAP on four image classification benchmarks. Finally, the concluding remarks and future work are reported in Section \ref{conclusion}.

%% file: sections/2_RelatedWork.tex
\section{Preliminaries: Considerations to generate Local Linear Explanations}
\label{background}

% En esta sección repasamos el tipo de explicaciones llamadas explicaciones lineales locales(LLE), tambien llamadas feature importance models, additive feature attribution methods or linear proxy models. Estos métodos son llamados LLE porque son una aproximación lineal local de la caja negra en el ejemplo $x$ que queremos explicar. 
% Un método de explicación de la black-box $f$ basada en LLE se construye como un modelo de white box $g$ que pretende imitar el comportamiento de $f$ en un entorno del ejemplo original $x$.

In this section we review the type of explanations named LLE, also called feature importance models, additive feature attribution methods or linear proxy models. These methods are called LLE because they are a local linear approximation of the black-box.

% Esta sección comienza con una descripción teórica sobre las LLE para, posteriormente, describir las dos LLEs del estado del arte, LIME y SHAP. Tras esto, discutimos cuatro aspectos fundamentales para la generación de feature importance explanations:  las diferencias entre el concepto de importancia y cómo compararlos, cómo generar el vecindario de ejemplos para la regresión de las LLEs, diferentes consideraciones sobre el tipo de dato en el que trabajemos y la tarea específica que abarcamos.

This section starts with a theoretical description of LLEs and describes the two state-of-the-art LLEs, LIME and SHAP. We then discuss four fundamental aspects for the generation of feature importance explanations: the differences between the concept of importance and how to compare them, how to generate the neighborhood of examples for the regression of LLEs and different considerations about the type of data we work on and the specific task we tackle.

\subsection{Local Linear Explanations}

%Let $x$ be an example of the data $\mathcal{X}$ which is fed to the black-box $f$. For the sake of generality, no assumptions on the details of $f$ are made. A possible XAI method of the black-box model $f$ based on LLE is a white-box model $g$ designed to mimic $f$ around $x$.

Formally, let $X \subset \mathbb{R}^F$ be the input dataset. Let $f:\mathbb{R}^F \rightarrow \mathbb{R}^C$ be the original black-box model, where $C$ is the dimension of the output space $\mathcal{Y}$. Previous works defines $f$ as a function that relies on just $\mathbb{R}$, but in case of tasks such as non-binary classification problems the model output is a vector of probabilities where each component depends on all others. Let $x \in  X$ be the input to be explained. A white-box LLE explainer is a function $g:\mathbb{R}^F \rightarrow \mathbb{R}^C$ defined as follows:

$$g(x) = Ax+B, A\in \mathcal{M}_{F,C},B \in \mathbb{R}^C,$$

in other words, $g$ is a linear application from the feature space to the output space. 

% De manera intuitiva, los pesos de la matriz A y el vector de sesgos B están fuertemente ligados con la importancia de cada característica. El peso $a_{i,j} de la matriz $A$ repercute directametne en la importanci de la característica $i$ para la salida $j$ y el peso $b_j$ está ligado a la importancia de la salida $j$ en general. 

Intuitively, the weights of both $A$ and $B$ are linked to the importance of each feature. 
More precisely, each weight $a_{i,j}$ of matrix $A$ is linked to the importance of feature $i$ to output $j$. 
Also, each bias $b_j$ is linked to the general importance of output $j$.

The different LLE methods use linear regression minimizing error as follows:

\begin{equation}
    \mathcal{L}(f,g,\pi_x) = \sum_{z \in N(x) }\pi_x(z)(f(z)-g(z))^2,
\end{equation}

where the weight function selection depend on each particular method.
Another factor to consider is how the neighbors are sampled. 
The original proposals consider a Bernouilli experiment for each feature, that is, each feature has the same probability to be present on the generated neighbour. 
On the other hand, there are other newer proposals that consider a smart perturbation generation \cite{slack2021reliable}, where examples that contribute more to the explainability white-box model are more likely to be generated.
For each LLE method, we use the sample-wise approach.

\subsection{Models of Local Linear Explanations: LIME \& SHAP}

% Una vez explicado lo que son las LLE, vamos a pasar a describir las dos LLE principales del estado del arte, Linear Model-agnostic Explanation (LIME) y SHapley Additive exPlanations (SHAP). Aunque ambas son LLEs, tienen claras diferencias a la hora de realizar la regresión de la caja negra. A continuación, pasamos a describir el funcionamiento de cada método y las diferencias principales entre ellos.

Once explained what LLEs are, we are going to describe the two main state-of-the-art LLEs, Linear Model-agnostic Explanation (LIME) and SHapley Additive exPlanations (SHAP). 
Although both are LLEs, they have clear differences in performing the black-box regression.
We now describe how each method works and the main differences between them.

\paragraph{\textbf{LIME}}

The LIME method \cite{ribeiro2016should} adopts the concept of local importance, which means that a feature that produces significant changes in the neighborhood of $x$ is very important. 
Therefore, features that are important for the classification of $x$ but do not produce significant changes in the neighborhood of $x$ will end up being discarded as an important feature. 

Formally, LIME build a LLE model $g$ by linear regression over a neighbourhood $N(x)$ of the original datapoint $x$. 
The definition of this neighborhood is not trivial due to each dataset's different nature.
In order to find a LLE $g$, LIME fits a Ridge regression to $N(x)$ with the linear least squares function with the default kernel:

\begin{equation}
    \pi_x(z) = exp(-d(x,z)^2/\sigma^2),
\end{equation}
where $d(\cdot,\cdot)$ is the euclidean distance and $\sigma$ is a regularization factor.

The generation of the neighbourhood $N(x)$ is performed by sampling from an exponential distribution with $\lambda=\dfrac{1}{\sigma}$ a value $v'$ with $\sigma$ the parameter selected for the LIME kernel. 
Finally, let $v=\lfloor v' \rfloor$. 
In the hypothetical case of $v > F + 1$, $v=F$ where $F$ is the number of all features. The value $v$ sampled is used to select randomly $v$ features to exclude on this sample,

\paragraph{\textbf{SHAP}}

The SHAP method \cite{NIPS2017_7062} considers a feature to be important for the classification of an example $x$ if it produces significant changes when compared to background values.

Formally, SHAP build a LLE model $g$ by computing the contribution of each feature to the prediction from a game theory approximation. 
This method tries to find a LLE $g$ as a regression with the following kernel function, which is the SHAP kernel $\pi_x$ defined as follows:

\begin{equation}
    \pi_x(z) = \frac{F-1}{{\binom{F}{|z|}}(F-|z|)|z|},
\end{equation}

where $z \in \{0,1\}^F$ is a binary vector representing the presence of each of the $F$ features on the $z$ example and $\binom{N}{M}$ is the combinatory number of choosing $M$ elements from $N$ possibilities without replacement.

This method can obtain an exact explaination $g$ if we evaluate all the possible examples of $z$, that is, $2^F$ evaluations of the black-box $f$. As the number of evaluations required increases factorially with respect to the number of features, this non-stochastic approximation is unaffordable. That is why the general use of this method uses also an stochastic approximation generating a list of $N$ different examples and solve the linear Ridge regression as LIME does.

The generation of the neighbourhood $N(x)$ is performed by sampling a value $v$ from a random discrete variable $X$ whom distribution is the following:

\begin{equation}
    P[X=x]=\dfrac{\dfrac{1}{(x+1)(M-x+1)}}{\sum_{i=0}^{M}\dfrac{1}{(i+1)(M-i+1)} },x=0,...,M
\end{equation}
    
that is, the random variable $v$ that assign to $i$ the proportional probability of the weight that SHAP assigns to all the instances that excludes exactly $i$ variables. 
The value $v$ sampled is used to select randomly $v$ features to exclude on this sample.

\subsection{How to define features for LLE in non-tabular data}%Feature definition in non-tabular data for LLE/ How to define features for LLE in non-tabular data

% Tipo de dato de imágen => dato binario
% Para una explicación basada en feature inportance, es muy importante definir qué va a ser una característica. En datos tabulares, la definición de característica se deriva de forma natural del propio dataset. Sin embargo, otros tipos de dato no tienen esa facilidad, por ejemplo, las series temporales o las imágenes. En el caso de las series temporales, la mínima cantidad de información la obtenemos de cada momento de medición. En el caso de imágenes, la obtenemos del pixel. Esto tiene dos problemas asociados:

For an explanation based on feature importance, it is very important to define what a feature is. In tabular data, a feature is defined naturally from the dataset itself.  However, other types of data do not have this convenience, e.g., time series or images. In the case of time series, the minimum amount of information is obtained at each measurement timestep. In the case of images, we get it from each pixel. This has several associated problems:

% - \textbf{Los modelos de explicabilidad exactos son inabordables.} En el caso de SHAP, para una cantidad de $F$ features se necesitan $2^F$ evaluaciones de la black-box para desarrollar una explicación. 
% - \textbf{Las explicaciones pierden perspectiva.} El valor de un único dato no suele aportar demasiada información sobre series temporales o imágenes. Sin embargo, una agrupación de datos aporta más información. 

\begin{itemize}
    \item \textbf{Generating exact explanations becomes an unaffordable task.} In the case of SHAP, for a number of $F$ features, $2^F$ evaluations of the black-box are needed to generate the non-probabilistic explanation. A generic imagenet image has a size of $224\cdot224=50176$ pixels, resulting in $2^{50176}$ black-box evaluations in SHAP. Even in its probabilistic versions, a regression needs a large number of these evaluations to be reliable.
    \item \textbf{Explanations loose perspective}. For a human being, a single pixel means nothing. In order to make a meaningful explanation, several pixels must be grouped together.
\end{itemize}

To solve these problems, some works use a division of the image into squares of the same size \cite{zhu2019guideline} while others use an unsupervised segmentation method to generate larger segment-size features \cite{schallner2019effect}.

% Ideas:
% Intro. La explicabilidad en imágenes con model agnostic
%   - Pixel a pixel inabordable
%   - Tapar regiones de la imagen
%   - Superpixel vs recuadro(optamos recuadro)

\subsection{How to explain with LLE in different Machine Learning tasks}%How to explain with LLE in different ML tasks
\label{task-specific}

To explain an artificial intelligence model, it is necessary to take into account the task for which the model has been designed. 

\begin{itemize}
    \item In the regression task, each element of the output can be explained separately. Thanks to this, no output is dependent on any other and a separate analysis can be performed. . 
    \item In classification task, the output is usually a vector of probabilities with clear constraints that must be satisfied (each element must be greater than or equal to 0 and the sum of all of them must be 1). Furthermore, it is not just the class to which it is classified that has an influence, but also the degree of certainty with which it is classified into each class. Since the outputs are dependent on each other in this case, a joint analysis of the output must be carried out.
    \item In the clustering task, an explanation can be carried out simply by some example or by some rule for each cluster \cite{9035478}. Therefore, it is necessary to unify the concept of explanation within the clustering task. 
\end{itemize}  

Therefore, for each specific task, a different method of explanation must be developed. From now on, we focus on the task of classification, described formally below.

\paragraph{\textbf{Classification task specifications}}Let $g$ be a local linear white-box-model where $g: F \xrightarrow{} Y_l $ over the logit space, $g(x) = Ax+B$. We define the signed importance matrix as the derivative matrix $A^l$, over the logits space. It should be noted that $A=A^l$.

To obtain the probability vector, we need to apply the softmax function, that is, $p= softmax(Ax + B)$.
We define $A^p = D(softmax(g(x)))(x)$ where $D()$ is the derivative operator. 

The component $a_{i,j}$ of matrix $A^l$ and $A^p$ will refer to the importance of feature $i$ for class $j$ over the logit and probability spaces respectively.

Both matrices give us important and complementary information about the behavior of the white-box $g$. The $A^l$ matrix gives us absolute information about how the logits of all classes respect to the original features. Additionally, the $A^p$ matrix gives us information about the classes that are potentially most likely to be classified as, disregarding the least likely. This may provide us apparently contradictory information, as we show in the following example:

\begin{itemize}
    \item Let $g:\mathcal{X} \xrightarrow{} \mathbb{R}^3$ the white-box linear model of a multiclass problem of three classes on the logit regresion and let $x$ be the original example. Let say $g(x) = (5,3,-2)$ and, therefore, $softmax(g(x))= (95.17\%,4.73\%,0.08\%)$. 
    \item We now consider $x'$, a neighbour of $x$ with a perturbation on $i$ feature, that produces $g(x')=(2.5,1.5,-1)$ and, therefore, $softmax(g(x')) = (71.52\%,26.31\%,2.15\%)$.
    \item If we consider exclusively the logit approximation, it may be interpreted as feature $i$ influences positively for classes 1 and 2 and negatively for class 3, with approximately the same intensity.
    \item If we consider exclusively the probability approximation, feature $i$ may has a positive influence for class 1, a negative influence for class 2 and, much less significantly, a negative influence for class 3.
\end{itemize}

From a global view-point, each view-point has its impact on the analysis. Thus, we define a new matrix $\mathcal{A}$ as the importance matrix and it is obtained as it follows from the matrices $A^l$ and $A^p$:

\[\mathcal{A}_{i,j} = sign(A^l_{i,j})\sqrt{|A^l_{i,j}| \cdot |A^p_{i,j}|},\]

that attempts to combine the information of both matrices $A^l$ and $A^p$. This matrix $\mathcal{A}$ has the sign of the logit matrix and the geometric mean of the intensity of importance of both matrices.

From the importance matrix $\mathcal{A}$, we define the relative importance matrix $\mathcal{A}_r$ as $\frac{\mathcal{A}}{\max_{a_{i,j} \in \mathcal{A}}(|a_{i,j}|)}$,the normalized matrix that maintains $0$ as $0$ and transforms the value with the greater absolute value to $1$ or $-1$, depending on the original sign of this specific value.

We define the absolute importance matrix $|\mathcal{A}|$ as the matrix of the terms $\mathcal{A}_r$ in absolute value, that is, $a_{i,j}= |a_{i,j}|$ for each coefficient $i,j$ of matrix $\mathcal{A}$. Each term $a_{i,j}$ of $|A_{i,j}|$ is the absolute importance of feature $i$ to the class $j$.

\subsection{Proposed frameworks to compute LLE: quantitative and qualitative approaches}%Proposed frameworks to compute LLE: quantitative and qualitative approaches

All proposed explanations are based on different notions of what constitute an explanation and, therefore, are not directly comparable. 
In the literature, there are several proposals to compare explanations.
In \cite{rosenfeld2021better} a set of metrics is proposed to measure the quality of explanations. However, they are specialized in rules-based explanations. 
In \cite{amparore2021trust}, the LEAF framework is proposed, with also four different metrics to evaluate agnostically different explanation metrics, independent of the explanation generation method. 
It also offers a practical example of their use, evaluating the quality of different explanations. However, the theoretical development of this framework is not mathematically consistent, which leads to biased conclusions.

It is in this scenario where the need for a mathematically consistent and unbiased explanation evaluation framework arises. In addition, this framework must also provide a measure not only comparative but also giving an absolute idea of the good behavior of the explanation itself.

%\subsection{Comparison between LIME and SHAP}

%Therefore, although both methods use the concept of importance to generate their explanations in principle, they are not comparable. 
%That is why it is of paramount importance to establish a set of metrics to evaluate different feature importance methods and compare them regardless of the concept of importance that each technique uses.

%% file: sections/3_Proposal.tex
\section{REVEL Framework}
\label{proposal}
% En esta sección comentamos diferentes aspectos teóricos de nuestro estudio.

%In this section we analyze the specifics of our problem, both at task level and at image data type level. We also propose metrics inspired by those analyzed in Section \ref{leaf}. The proposals has been made to be robust at a purely theoretical level.

In this Section, we propose a new explanation evaluation framework called REVEL Framework, presenting  five new metrics for assessing the quality of an explanation. In particular, for each metric proposed we describe the qualitative aspect  the metric is intended to measure and has guided its definition. We also provide a guideline on how to interpret the metric. Finally, for each qualitative aspect, we make a theoretical comparison of each metric with other proposed metrics.

On Table \ref{tab:metrics_summary}, we summarize the metrics we propose and the qualitative aspect they measure.

\begin{table}[H]
    \scriptsize
    \begin{tabularx}{\textwidth}{XX}\toprule

        \textbf{Name} &\textbf{What is evaluated} \\\cmidrule{1-2}
        \textbf{Local Concordance} &How similar is the LLE to the original black-box model on the original example \\\midrule
        \textbf{Local Fidelity} &How similar is the LLE to the original black-box model on a neighborhood of original example \\\midrule
        \textbf{Prescriptivity} &How similar is the LLE to the original black-box model on the closest neighbour that changes the class of the original example \\\midrule
        \textbf{Conciseness} &How brief and direct is the explanation \\\midrule
        \textbf{Robustness} &How much two explanations generated by the same LLE generator differ \\
        \bottomrule
    \end{tabularx}
    
    \caption{Summary of the metrics developed by REVEL and the qualitative aspect they measure}\label{tab:revel_metrics}
    \label{tab:metrics_summary}
\end{table}

\subsection{Explanation Local concordance}
\label{REVEL_Concordance}

There are LLE methods guaranteeing the white-box explanation and the black-box model to match on the specific datapoint. However, these methods have a strong computational constraint, since they require a large number of evaluations of the black-box model. Other methods do not ensure the coincidence between white-box explanation and black-box model. Since the concordance between both is not guaranteed, it is possible that the class proposed is different from each other, which means the proposed explanations end up being inconsistent. We want to measure how much the explanation and the model are similar.

On the classification task of more than two classes, it is also necessary to consider jointly the whole probability vector. Our proposal also attempts to measure the smoothness from the min to the max concordance values, that is, only the min concordance should have a score of 0 and the max concordance should have a score of 1 on this metric.

% Podemos abstraer esta función de pérdida que evalua nuestra métrica para que considere distancias vectoriales entre probabilidades:

We can easily abstract the loss function that evaluates our metric to consider vector distances among probability vectors:

\begin{equation}
    Local\_Concordance(g) = 1-\frac{|f(x)-g(x)|}{C},\label{eq:local_conc}
\end{equation}

where $|\cdot|$ is a defined norm (1-norm, 2-norm, inf-norm...) and $C$ is the maximum distance between two possible probability vectors. This term exists and is reached because the probability space is complete and the norm is continuous. Moreover, $C$ is computed as $|u-v|$, where $u=(1,0,...,0)$ and $v=(0,1,0,...,0)$, regardless of the norm.

This metric has the following qualities:
\begin{itemize}
    \item Using C as the normalization factor makes our score well defined in the interval $[0,1]$, with the max concordance achieving 1 and the min concordance achieving 0.
    \item This metric considers the whole probability vector jointly and not just one coordinate of the probability vector.
\end{itemize}

\paragraph{\textbf{Guideline}} This metric measures how similar the explanation is to the black-box in the original example. 
It is very important that this metric is close to 1. 
Otherwise, the proposed explanation does not explain what happens in the example itself.

\paragraph{\textbf{Comparison}} The analogous LEAF proposal local concordance is defined as $l( \left|f(x)-g(x)\right| )$, where $l(k) =max(0,1-k)$ is the Hinge loss function \cite{gentile1998linear}. In contrast to our proposal, the use of the Hinge function makes it non-smooth. It also does not assure that only the maximum discordance reaches the worst value of the metric. In conclusion, the LEAF proposal has inconsistencies that our proposal overcomes.

\subsection{Explanation Local Fidelity}
\label{REVEL_fidelity}
Local Fidelity applies not to a classification task but a regression one. 
The main idea of this metric is how close is the white-box $g$ approximating the probabilities obtained by the black-box $f$. 
We propose the mean concordance between probabilities of $g$ and $f$ obtained on the neighbourhood $N(x)$, that is,

\begin{equation}
    Local\_Fidelity(g) = \frac{1}{|N(x)|}\sum_{n \in N(x)} 1-\frac{|f(n)-g(n)|}{C}.\label{eq:local_fifel}
\end{equation}

This metric is an extension of the local concordance on $x$ extended to its neighbourhood $N(x)$. It is also well-defined on the interval $[0,1]$.

\paragraph{\textbf{Guideline}} This metric measures the similarity between the explanation and the black-box in the neighborhood.
This metric is essential to check that the tendency of the explanation is similar to the tendency of the black-box. 
It must be close to 1 to obtain a good explanation.

\paragraph{\textbf{Comparison}} The analogous LEAF metric proposes to evaluate the resemblance between the white-box explanation and the black-box model in the proposed neighborhood N(x) using the F1 metric.

\begin{itemize}
    % Esta métrica está especializada en problemas de clasificación. Puesto que $N(x)$ es un vecindario de $x$, la mayoría de los ejemplos serán, por continuidad, de la misma clase que x, provocando un desbalanceo en el conjunto N(x). 
    \item The LEAF proposal is a measure designed to evaluate classification problems. Since $N(x)$ is a neighborhood of $x$, most examples will, by continuity, be of the same class as x, resulting in an imbalance in N(x).

    %en la frontera de decisión a la hora de medir similitud. Sea $x'$ un ejemplo del conjunto $N(x)$ donde $g(x')=0.49$ y $f(x') = 0.51$ para un problema binario. La métrica F1 penalizará mucho este ejemplo mientras que la realidad es que $f$ y $g$ son muy similares en este ejemplo $x'$.
    \item This metric presents problems at decision borders. In a binary problem  with threshold $0.5$, let $x'$ be an example of set $N(x)$ where $g(x')=0.49$ and $f(x') = 0.51$. The F1 metric will penalize this example while actually the white-box $g$ mimics almost perfectly the undecidability of the black-box $f$.
\end{itemize}

Our proposal has no problem with the imbalance dataset generated by $N(x)$ for the metric evaluation. Also, our metric is not biased by a threshold selection.

\subsection{Explanation Prescriptivity}

The main idea of prescriptivity is to test whether the white-box explanation $g$ has correctly predicted the changes needed in the original example in order to change the original class.

Mathematically, let $x$ be the original example, $f$ the black-box model, $g$ the white box model mimicking $f$ and $h$ the changes needed on $x$ to change the class predicted by the white box $g$. We propose the following prescriptivity metric:

\begin{equation}
    Prescriptivity(g) = 1-\frac{||f(x+h)-g(x+h)||}{C},\label{eq:prescriptivity}
\end{equation}

where $C$ is a normalization factor. This normalization factor is the same as in equation \ref{eq:local_conc}.

In our proposal, $h$ is obtained by removing the presence of the most important positive features of the  class predicted by the white-box $g$ on the example $x$. The algorithm ends when $g$ assigns a different class to $x$ and $x+h$, that is, $argmax(g(x)) \neq argmax(g(x+h))$.

This metric has the following properties:

\begin{itemize}
    \item This prescriptivity proposal is defined as a vectorized proposal so the metric has a global view of the whole output. 
    \item This metric obtains the maximum value $1$ when both vectors $u$ and $v$ are equal and obtain the minimum value $0$ when both vectors are in the maximum possible disagreement on this prescriptivity scenario.
    \item This metric is not dependant of a boundary selection. Neither it is dependant on a specified neighbourhood $N(x)$.
\end{itemize}

\paragraph{\textbf{Guideline}} 
Prescriptivity challenges the explanation to propose an example far enough to change the prediction of the model but without losing predictive quality at this point.
Indirectly, each explanation proposes an example $x'$ different from the original example $x$ and whose prediction must be markedly different from that of $x$.  
Although the best possible score for this metric is 1, it is understandable that it does not reach the best score and serves more as a comparative metric between different explanations method.

\paragraph{\textbf{Comparison with LEAF}}

The prescriptivity metric is formally proposed in LEAF for a binary classification problem, where a fixed decision boundary is chosen. This decision boundary is the set $\mathcal{D}_g(y') =\{ x \in \mathbb{R}_F : g(x) =y'\}$, that is, the set of points in the domain whose prediction by the white-box is exactly $y'$. 

On the LEAF proposal, the obtention of $h$ is based on the closest projection of our example $x$ on $\mathcal{D}_g(y')$. In reality, this is only possible if the features selected are real-valued. In case of binary data, this approximation can't be achieved because each feature can't process a real-value. It is also dependent on a selection of a boundary $y'$.

LEAF proposes as prescritivity metric the following function:

\begin{equation}
    l\left(  \frac{ \lvert f(x') - g(x') \rvert }{C} \right) ,
\end{equation}

where $l(\cdot)$ is the hinge loss function, and $C=max(y',1 - y')$ is a normalisation factor, so that 1 means that $x'$ lies at the boundary, and 0 means x' is at the furthest distance from the boundary. One may observe that by taking the absolute value, the measure both over-shoot and under-shoot the boundary as a loss of prescriptivity.

The LEAF proposal has different problems:

\begin{itemize}
      
    \item This metric is designed for a single output variable. For classification problems, it is usual to obtain a vector of probabilities whose components are linked to each other and whose analysis must be done jointly. 
    \item Choosing a fixed $y'$ value does not guarantee the change of class when we talk about non-binary classification problems. In case of a classification problem of more than two classes, the majority class could have a 50\% probability and other classes could share the rest of the probability equally. This result on a $x'$ neighbour of $x$ whose changes does not change the original class.
    \item The proposed norm is restricted to the interval $[0,1]$ but not smothly. Even if it is used a normalization parameter $C$, it is not clear if only the maximum possible disagreement results in a 0 score on this metric or if it is even reachable. It is reasonable for this kind of metric to guarantee that the maximum disagreement obtains 0 as the worst score and, as agreement increases, the metric increases smoothly up to 1, the maximum score.
\end{itemize}

Our proposal does not show all of the different problems detected in the LEAF prescriptivity proposal, since our metric jointly measures the full probability vector, is not boundary dependent and is well defined in the interval [0,1], where it smoothly where it changes smoothly from worst case to the best one.

\subsection{Explanation Conciseness}

% Como vimos en \label{concissness}, la propuesta que hace LEAF es una restricción a los métodos y no una métrica real. En consecuencia, proponemos la sigueinte métrica basada en la matriz de importancia absoluta $\mathcal{B}$$
Conciseness measure aims to evaluate the brevity of the explanation. In our case, the less relevant features our explanation has, the more concise it should be. 

We propose the following conciseness metric based on the absolute importance matrix $|\mathcal{A}|$, particularly in the vectors of importance of each feature. Let $v_i = (a_{i,1},...,a_{i,N})$ be the importance vector of feature $i$, where the coefficient $a_{i,j}$ is the $i,j$ coefficient of matrix $|\mathcal{A}|$. We define the conciseness of the explanation proposed by the white-box $g$ as

$$ Conciseness(g) = \frac{1}{f-1}\sum_{i=1}^{f}1-|v_{i}|_1.$$

which can be described as the mean irrelevance of the features. If we consider $|v_{i}|$ instead of $1-|v_{i}|_1$ we would have the mean relevance of the features and the most concise method would have a score of $\dfrac{1}{f-1}$. That is why we have reversed this term.

This metric has the following qualities:

\begin{itemize}
    \item It rewards the use of few features with a high weight.
    \item We have a general idea of how many features are important on the white box.
    \item The best possible score is obtained if we have only one feature with absolute importance 1 and the rest  with 0 absolute importance, in which case we would obtain $1$ as conciseness. The worst case it is obtained when we have all the features with $1$ as absolute importance, in which case we would obtain $0$ as concissness.
    \item We can compare explanations with different amount of features taken into account.
    
\end{itemize}

\paragraph{\textbf{Guideline}} This metric evaluates the ability of the explanation to focus on the most important features of an example and discard the less important ones.
%A score of 1 means that only one of the features is considered important while a score of 0 indicates that all features are equally important. 
Depending on the complexity of the explanation we want, we may prefer greater or lesser conciseness. 
For instance, in image classification the explanation to dismiss a large part of the image could be desired but not to have a single pixel explaining the complete decision of the model.

\paragraph{\textbf{Comparison}} 

LEAF proposes as conciseness a constraint for explanations, where it requires that explanations use exclusively $k$ features. In the case of LIME, conciseness is a variable that we supply to the algorithm so that it restricts itself to choose a given number of features with non-zero importance. On the other hand, in the case of SHAP, the algorithm uses by default all available features and gives them an importance. In order to compare both methods, the LEAF framework proposes to select a default conciseness parameter k the number of features to be used on the white-box explanation and restrict both LIME and SHAP to use the top-k most important features.

As mentioned in the previous paragraph, the proposed conciseness is not a metric but a constraint on white-box explanation models. Moreover, the LEAF proposal do not leave the white-box models decide whether a particular decision has been influenced by more or fewer features.

Our proposal, instead of a constraint, provides a metric to evaluate the conciseness of each white-box explanation.

\subsection{Robustness over explanations}

A key point to consider is the variability of the methods used to generate explanations. It is desirable that independent explanations generated by the same method must be as similar as possible, since very different or even contradictory explanations would lead to mistrust the method. In case of deterministic methods, this is ensured since there is just one proposed explanation. In case of non-deterministic methods, there are several proposed explanations and, therefore, we need to ensure that the explanations does not differ or even contradict each other.

To measure how two explanations $g$ and $g'$ differ we propose two possible measures:

\begin{itemize}
    \item First, we propose the cosine similarity between $\mathcal{A}_r$ and $\mathcal{A}_r'$, which are the relative importance matrices of $g$ and $g'$ respectively: 

    $$ sim\_cos(g,g')= \frac{\mathcal{A}_r\cdot \mathcal{A}_{r}'}{|\mathcal{A}_r||\mathcal{A}_{r}'|},$$
    
    where $\cdot$ is the scalar product.
    
    \item The metric proposed before based on the cosine similarity take into account the direction of the matrices $\mathcal{A}$ and $\mathcal{A}'$  but not the magnitude. To take the magnitude also into account, we propose the following measure of similarity:
    
     $$ similarity'(g,g')= \left(  \frac{\mathcal{A}_r\cdot \mathcal{A}_{r}'}{|\mathcal{A}_r||\mathcal{A}_{r}'|}\right)\left(1- \left|\frac{ ||\mathcal{A}_r||-||\mathcal{A}_{r}'||}{\max(||\mathcal{A}_r||,||\mathcal{A}_{r}'||)}\right|\right),$$
    
    that takes into account both, direction and magnitude of the explanations. In case of the same magnitude, this similarity function is exactly the cosine similarity. In case of different magnitude, this similarity function has lesser punctuation than the cosine similarity in case of a positive scalar product. In case of a negative scalar product, this score has also a lesser absolute value than the cosine similarity. In case of perpendicular explanation vectors, both metrics have a $0$ score.

\end{itemize}

In both cases, as robustness we propose the mathematical expectation of the chosen similarity of two different explanations $g$ and $g'$, that is:

$$ robustness(G)= \mathbb{E}[similarity(g,g')],$$

where $G$ is the set of all explanations that could be proposed by a certain explanation method such as LIME or SHAP. The expectation can be approximated by generating a given number of explanations and computing the mean of the similarities among explanations.

Those metrics have the following qualities:

\begin{itemize}
    \item Both metrics take into account the weight of all features, so two explanations $g$ and $g'$ choosing a different most-important feature  would be punished by both metrics.
    \item The second metric takes into account the magnitude of the importance  matrix.
\end{itemize}

\paragraph{\textbf{Guideline}} This metric does not evaluate a specific explanation but the method that generates them.
%It evaluates how similar two explanations generated by the same method tend to be. 
All deterministic methods will score 1 in this metric since they always generate the same explanation. 
Therefore, this metric is designed to evaluate the robustness of non-deterministic methods.
The closer this metric is to 1, the less the explanations generated by this method vary. 
It should be noted that this metric, due to the way it is designed, can give negative scores, which would indicate that the proposed explanations are contradictory.

\paragraph{\textbf{Comparison}} LEAF proposes the reiteration similarity metric, which measures how much two explanations generated by the same method vary by measuring the difference between the top-k features over several explanations proposal.

\begin{itemize}
    \item This metric depends directly on the conciseness constraint of the LEAF proposal.
    \item This metric does not consider the importance of a feature, since it penalizes equally for choosing important and not so important features, not penalizing it.
    \item This metric does not penalize choosing a "positive" important feature as "negative" and viceversa. Two different explanations can consider using the same feature $i$ for their explanation but attributing positive importance to it in the first explanation and negative importance in the second, which is a clear contradiction. The similarity proposal do not see this example as a contradiction and do not penalize it.
\end{itemize}

Our proposed robustness metric does not depend on external constraints and does not have the shortcomings described above while still measuring the variation between generated explanations.

%% file: sections/4_ExperimentalSetup.tex
\section{Experimental setup}
\label{experimental_setup}
In this section we describe the experimental setup we use in this work. We first select four image datasets as benchmark where we train the models to explain. Finally, we fix some hyper parameters to compare different LLE aspects with the REVEL framework.

\subsection{Benchmark selection}

The datasets selected as benchmarks are CIFAR10 \cite{cifar10}, CIFAR100 \cite{cifar100}, FashionMNIST \cite{xiao_fashion-mnist_2017} and EMNIST balanced \cite{cohen2017emnist}, which is a benchmark already used in \cite{yan2022towards} for explainability tasks.. Table \ref{Datasets} shows a short description of each dataset.

\begin{table}[http]
    \label{Datasets}
    \centering
    \caption{Descriptive table of the benchmarks selected}
    \label{tab:my_label}
    \begin{tabular}{cccccc}
        \toprule
        Dataset & Nº classes & Original image size & Training & Test& RGB  \\ \hline
        CIFAR10 & 10 & $32\cdot32$ & 50.000 & 10.000 &Yes \\
        CIFAR100 & 100 & $32\cdot32$ & 50.000 & 10.000 & Yes \\
        FashionMNIST & 10 & $28\cdot28$ & 60.000 & 10.000 & No \\
        EMNIST-balanced & 47 & $28\cdot28$ & 112.800 & 18.800  & No\\
    \end{tabular}
    
\end{table}

\subsection{General training pipeline}

For this experiment, we chose the efficientnet-b2 model \cite{tan2019efficientnet} with the pre-trained weights in the Imagenet dataset. 
Next, the network has been fine tuned on the benchmark dataset for 100 epochs, 32 images per batch with the Adam optimizer \cite{kingma2014adam} with learning-rate 1e-5, weight-decay=0.001 and amsgrad=True. We randomly selected a 10\% of the training set as validation subset on which the loss is not computed. Over the 100 epochs  models, we select the model which performance on this validation subset is the best.  As the objective of this work is the analysis of the metrics behavior, we will not go deeper into the training of the network and we will set these parameters as default. On Table \ref{tab:model-results} we show the performance obtained by the model in the different test sets of the datasets used as benchmarks.

\begin{table}[t]
    \centering
    \footnotesize
    \caption{Performance of the default model on the test sets of the selected datasets}

    \label{tab:model-results}
    \begin{tabular}{ccc}
    \toprule
        Dataset & Train/Test Partition & Classification Model Top-1 Accuracy (Test) \\ \hline
        CIFAR10 & 83.3\%/16.7\% & 95.26\%\\
        CIFAR100 & 83.3\%/16.7\% & 81.84\% \\
        FashionMNIST & 86\%/14\%& 94.25\% \\
        EMNIST & 86\%/14\% & 90.66\%\\
    \end{tabular}
    
\end{table}

\subsection{Local Linear Explanation pipeline}

On this subsection, with the purpose of generate a fair comparison, we fix as default some shared hyper-parameters of the LLE generation models, explained below.

\textbf{Number of neighbours (N):} For each example of the test split, we will generate a different number of neighbours examples to explain the original example. On the experiments, $N=100,200,300,400,500,600,700,800$.

\textbf{Neighbours generation (N(x)):} We use a smart perturbation generator, where each neighbour is generated with a probability proportional to the weight associated to it in each method of explanation generation.

\textbf{Number of explanations generated(E):} For each LLE method and each instance to be explained, we will generate \textbf{5} different explanations.

\textbf{Number of features of each image($F\cdot F$):} We divide each image into square patches of size $\dfrac{224}{F}\cdot\dfrac{224}{F}$, so each image will have $F\cdot F$ features.

\textbf{Feature occlusion:} To set a feature as occluded, we set the original patch from its original value to a neutral grey patch, that is, we set all pixel of the patch to 0.5 on each RGB channel. 

\subsection{On the comparison between LEAF and REVEL}

This paper presents REVEL as a proposal of theoretically robust measures for the evaluation of LLE explanations. The comparison with other measurement proposals, such as LEAF, should be carried out theoretically and not practically, since the measurements offered by the different proposals have nothing related to each other. That is why the comparison on this work is made exclusively on the theoretical proposal and not on the practical use cases.

%% file: sections/5_Experiments.tex
\section{Assessing explanations using REVEL: Use cases}
\label{experiments}
% En esta sección analizamos los resultados obtenidos para cada métrica por los diferentes generadores de explicaciones. Para ello, mostramos una gráfica de violín que muestra la distribución de las distintas puntuaciones obtenidas por los generadores de explicaciones.

In this section we propose three different scenarios in which REVEL can be used, thus demonstrating its analytical potential. These scenarios are:

\begin{itemize}

    \item \textbf{Dependence of LIME on the number of features (Section \ref{analisisFeatures}):} In this scenario we study how much the number of patches into which we have divided the original image can influence, or if there is an ideal partition in which to divide the images. 
    
    \item \textbf{Dependence of LIME on the number of black box evaluations (Section \ref{analisisblackbox}):} In this scenario we analyze the number of black-box evaluations needed to generate a good-quality explanation. We also evaluate the trade off between quality and time needed to generate a good explanation.
    
    \item \textbf{LIME vs SHAP (Section \ref{analisisLIMEvsSHAP}):} We compare the results obtained by the two state-of-the-art explanation generator models, LIME and SHAP, with the best configuration determined by the above scenarios. This scenario provides an idea about which explanation generator can offer us better explanations depending on their scores in each of the proposed metrics. 
\end{itemize}

\subsection{Dependence of LIME on the number of features}
\label{analisisFeatures}

% Esta comparativa nos sirve para realizar tanto un estudio general como un estudio centrándonos en el tipo de dato imágen. A nivel general, podemos estudiar de qué manera influye el número de características a la bondad de la explicación. Para el caso de imágenes, podemos utilizar este estudio para determinar la granularidad que mejor comportamiento tenga.

In this section we compare how LIME performs over different number of features. This comparison allows to perform both a general study and a study focusing on the image data type. At a general level, we analyze how the number of features influences the quality of the explanation. In the case of images, we use this study to determine the best performing granularity.

\begin{figure}[t]
    \centering
    \includegraphics[height=0.5\textwidth]{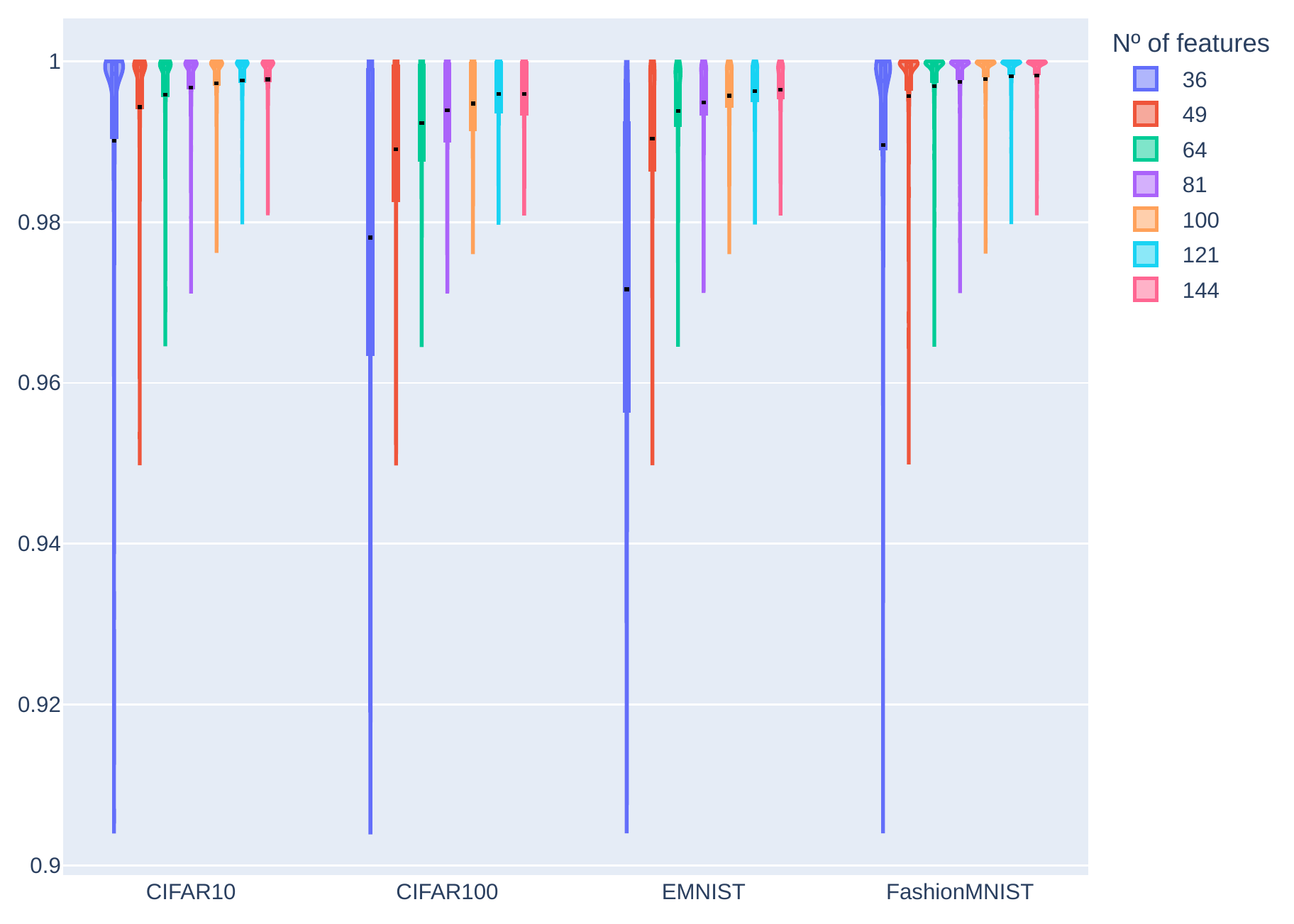}
    \caption{Performance of LIME methods grouped over number of features used on the \textbf{Local Concordance} metric.}
    \label{fig:concordance_NF}
\end{figure}

\paragraph{\textbf{Local Concordance}} In Figure \ref{fig:concordance_NF} we note that, as a tendency, the Local Concordance score increases the more features are processed. As the number of features increases, the explanation method has more parameters to fit. Therefore, the model increase its performance on mimicking the black-box on the original example.

% Notamos que, por norma general, la puntuación de Local Concordance aumenta cuantas más características se procesan. Esto es natural ya que con el aumento de número de características el método de explicabilidad tiene más variables que ajustar. Debido a ello, la consecuencia lógca es que el modelo se acerque mucho al ejemplo original.

\begin{figure}[t]
    \centering
    \includegraphics[height=0.5\textwidth]{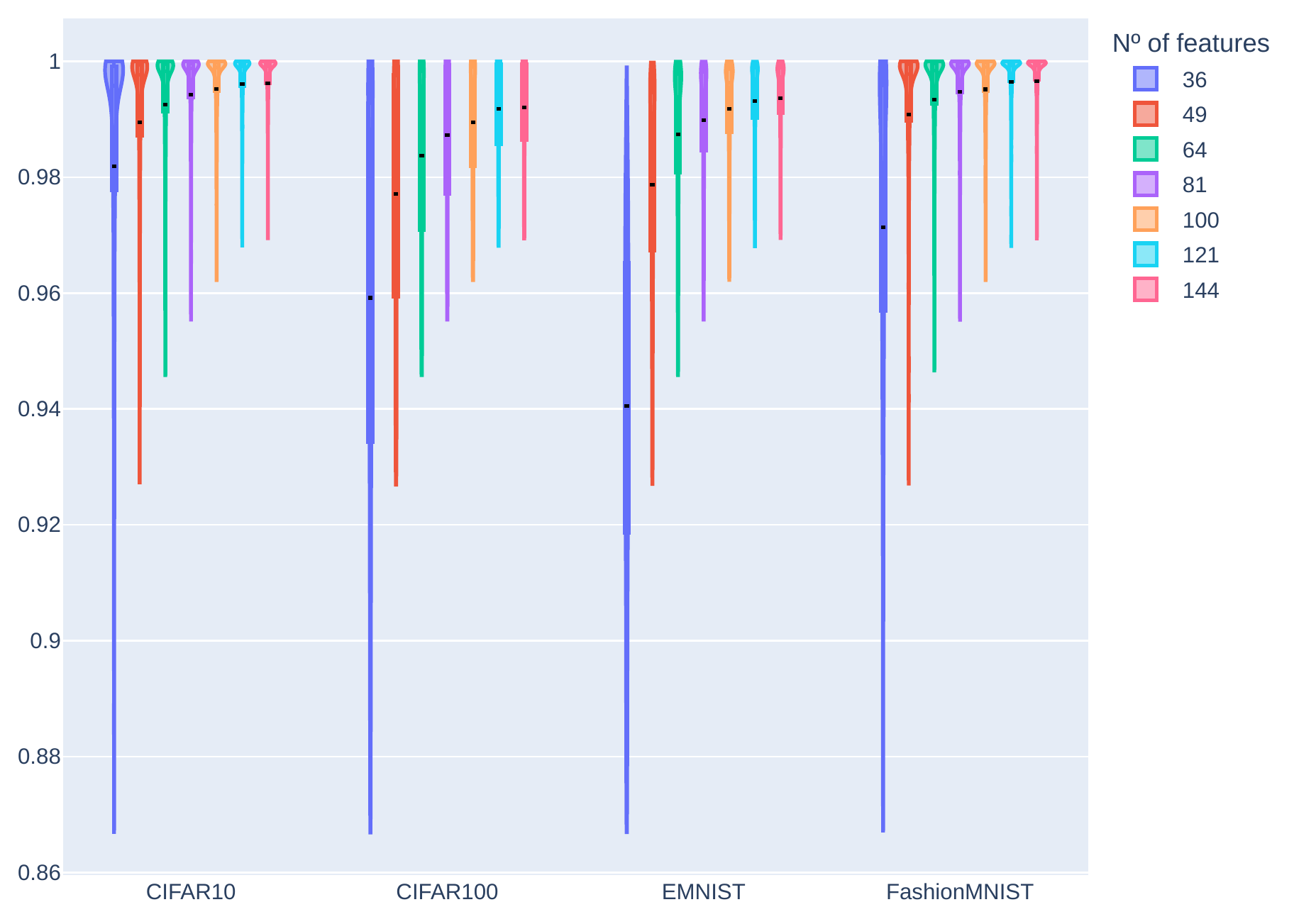}
    \caption{Performance of LIME methods grouped over number of features used in the \textbf{Local Concordance} metric.}
    \label{fig:fidelity_NF}
\end{figure}

\begin{figure}[t]
    \centering
    \includegraphics[height=0.5\textwidth]{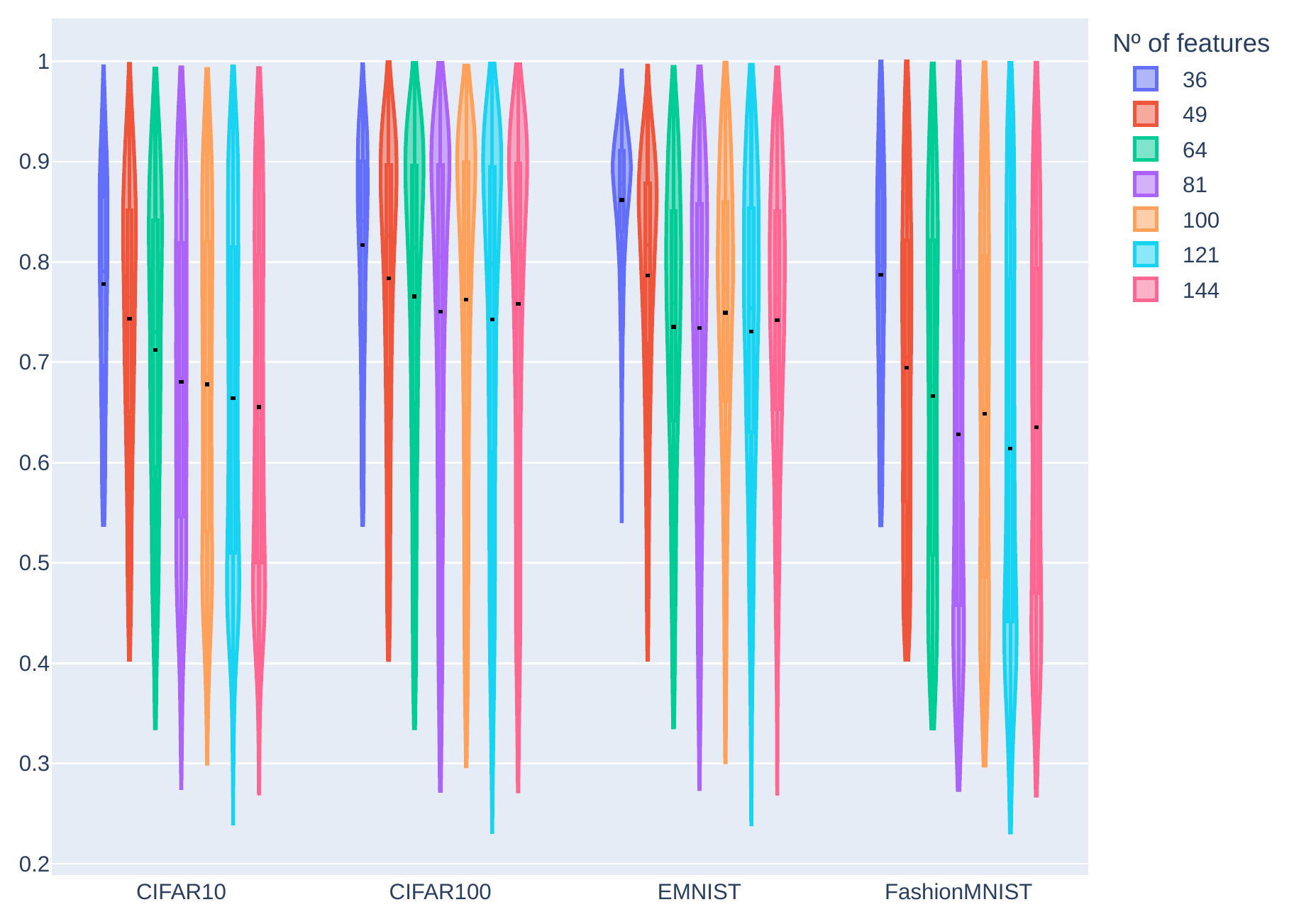}
    \caption{Performance of LIME methods grouped over number of features used in the \textbf{Prescriptivity} metric.}
    \label{fig:prescriptivity_NF}
\end{figure}

\paragraph{\textbf{Local Fidelity}}

In Figure \ref{fig:fidelity_NF}, we note a tendency similar to the Local Concordance. That is, Local Fidelity increases the more features we use. This is natural since the neighbors where we are evaluating Local Fidelity are closer to the original example the more features we use.

% Notamos una tendencia similar a la que existe con la Local Concordance. Esto es, aumenta la fidelidad local cuanto más características utilizamos. Esto es natural ya que los vecinos donde estamos evaluando la fidelidad local están más cerca cuanto más características utilizamos.

\paragraph{\textbf{Prescriptivity}}

On Figure \ref{fig:prescriptivity_NF}, in contrast to the Local Concordance and Local Fidelity metrics, a different pattern arises, where as the number of features increases, the Prescriptivity metric gets worse. Prescriptivity not only evaluates how well the explanation mimics the black-box in areas near the original example but also evaluates the proposed changes to the white box. The fewer features considered in the explanation, the fewer changes are necessary to change the predicted class. Thus, the explanation has less problems in finding the necessary features for the class to change.

% Al contrario que en las métricas de Local Concordance y Local Fidelity, notamos una tendencia muy distinta, donde al aumentar el número de características la métrica de prescriptividad empeora. Esto se debe a que la prescriptividad no solo evalua cómo de bien imita la explicación a la caja negra en zonas muy cercanas al ejemplo original sino que evalua la propuesta de cambios de la caja blanca. Cuantas menos características se consideren en la explicación, menos cambios serán necesarios para cambiar la clase predicha. Es por ello que a la explicación le cuesta menos encontrar las características necesarias para el cambio de clase.

\begin{figure}[t]
    \centering
    \includegraphics[height=0.5\textwidth]{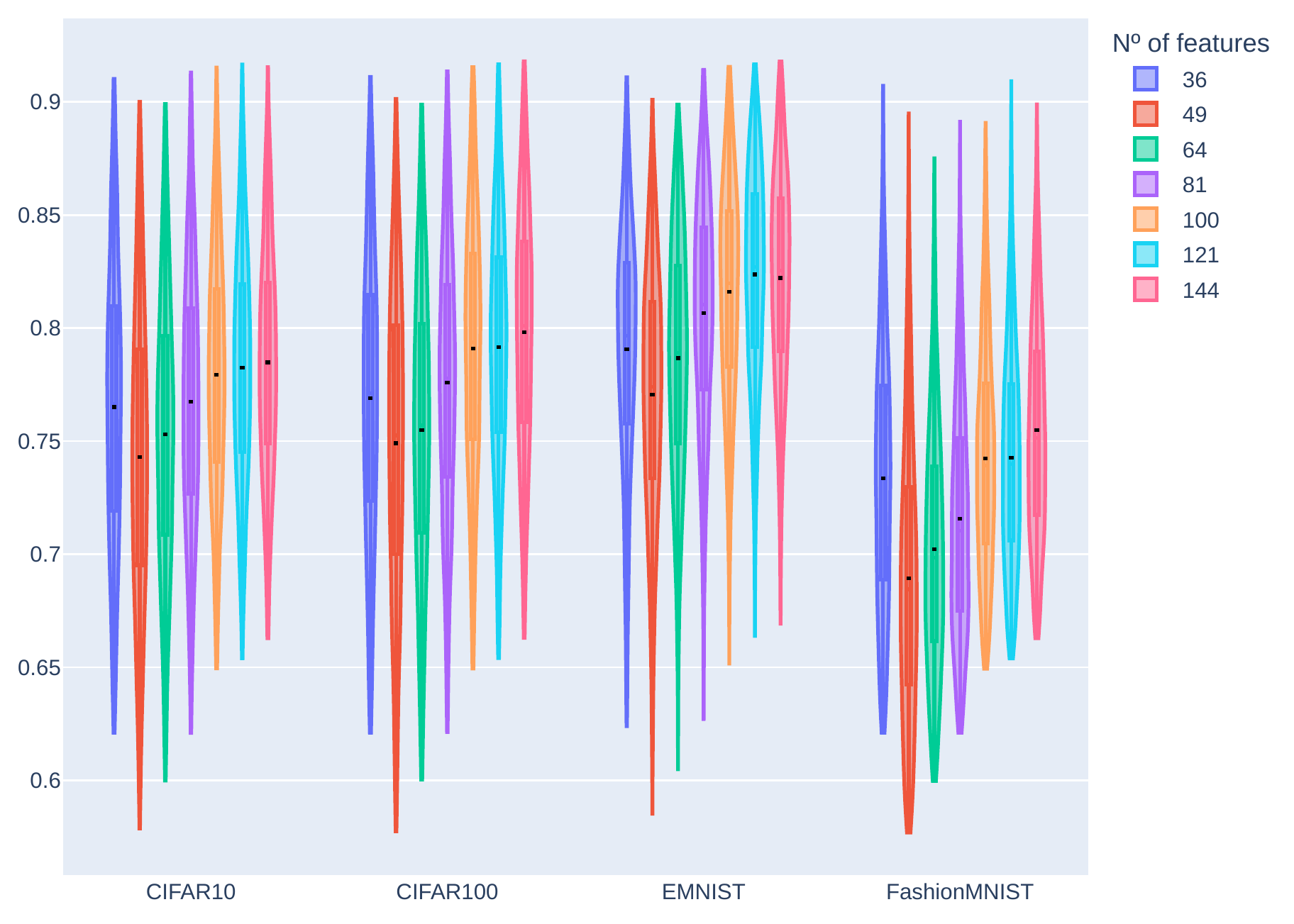}
    \caption{Performance of LIME methods grouped over number of features used in the \textbf{Conciseness} metric.}
    \label{fig:conciseness_NF}
\end{figure}

\begin{figure}[t]
    \centering
    \includegraphics[height=0.5\textwidth]{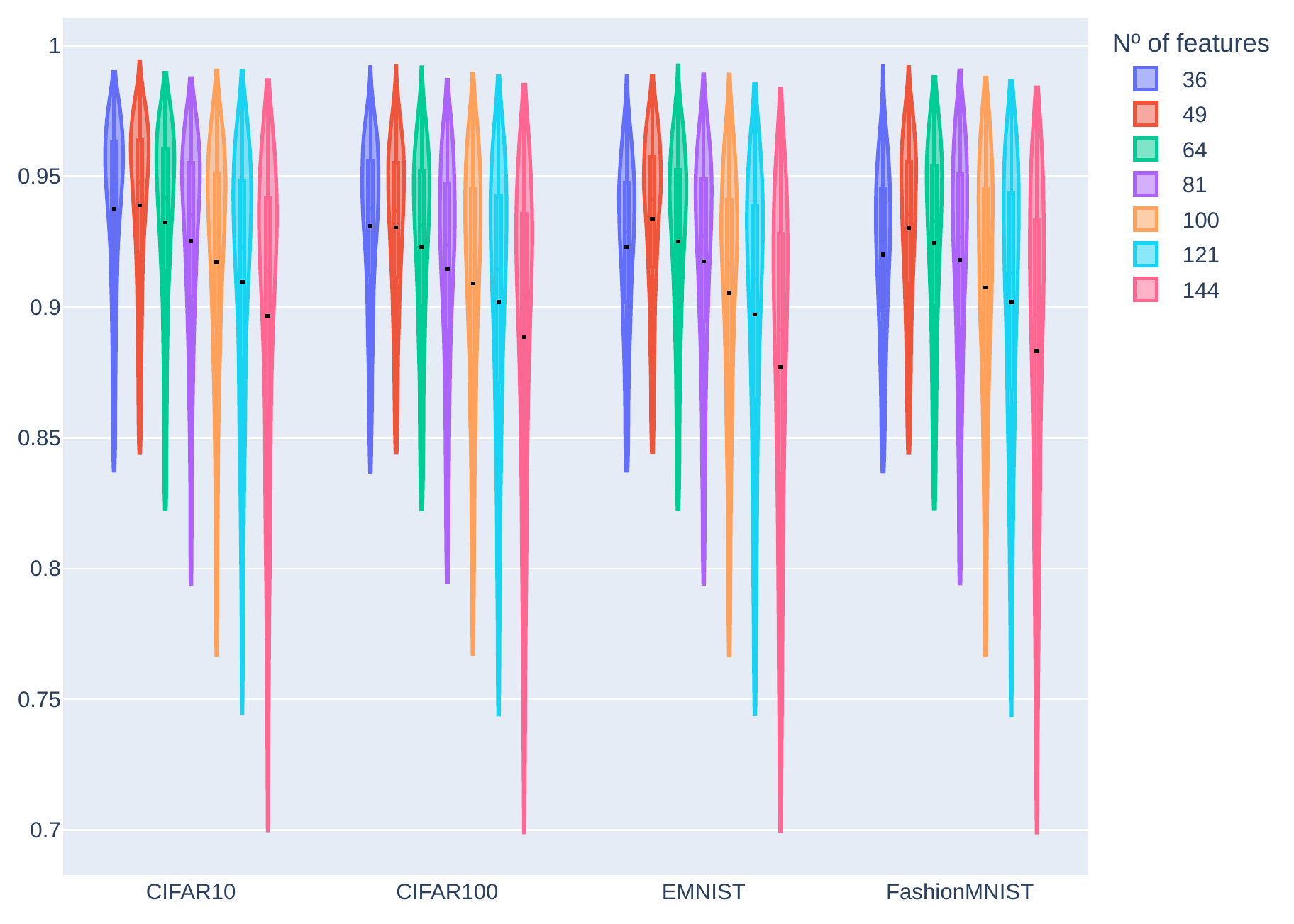}
    \caption{Performance of LIME methods grouped over number of features used in the \textbf{Robustness} metric.}
    \label{fig:reiteration_NF}
\end{figure}

\begin{figure}[t]
    \centering
    \includegraphics[height=0.5\textwidth]{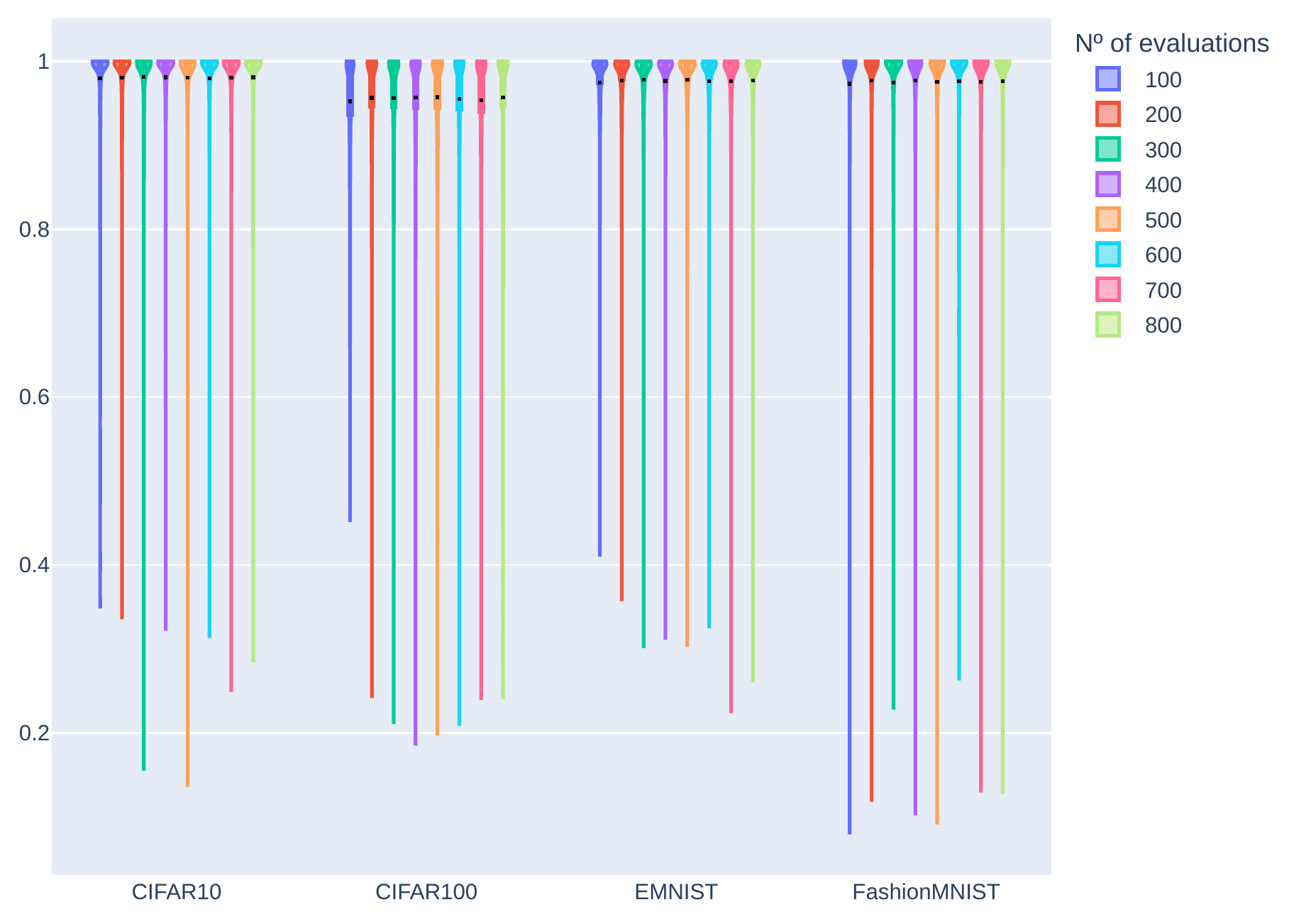}
    \caption{Performance of LIME methods grouped over max number of evaluations in the local \textbf{concordance} metric.}
    \label{fig:concordance_ME}
\end{figure}

\paragraph{\textbf{Conciseness}}

% Notamos que existe una tendencia a aumentar la conciseness conforme se aumenta la granularidad. Sin embargo, vemos que antes de este aumento, esta concisión decrece con 64 features. Esto parece indicar que cuando el número de features aumenta, mejor comportamiento obtendremos. Sin embargo, se puede interpretar como un sobreajuste de la explicación y que la mínima cantidad de información que se puede obtener de la imagen es partiendola en 64 características diferentes y que realizar una granularidad mayor sobreajusta al modelo. Aun así, habría que realizar un estudio con imágenes con diversos tamaños.

In Figure \ref{fig:conciseness_NF}, we note a tendency to increase Conciseness as the granularity increases. However, we observe that before this increase, Conciseness decreases with 64 features. This seems to indicate that the higher the number of features, the better the performance. However, it can also be interpreted as an overfitting of the explanation and that the minimum amount of information that can be obtained from the image is by separating it into 64 different features and that a higher granularity overfits the model. Even so, a study with images of various resolutions should be done because it could depend on the information contained on each patch.

\paragraph{\textbf{Robustness}}

In Figure \ref{fig:reiteration_NF}, we observe that the more features the models use, the more unstable the method becomes. Having more features to evaluate leads to more uncertainty in the choice of explanations.

% Podemos apreciar que cuantas más características utilizan los modelos, más inestable se vuelve el método. Esto se debe a que tener más características que evaluar propicia que las explicaciones tengan más incertidumbre a la hora de escogerlas.

\paragraph{\textbf{Global conclusion}}

We appreciate that the higher the number of features, the better the local performance. This is an expected result since is biased by the neighborhood we have chosen to calculate the Local Fidelity. Therefore, we should focus on the rest of the metrics. In the Prescriptivity calculation we see that the more features, the worse result is obtained. In contrast, the more features we see the more concise the methods are, discarding more unimportant features. Finally, we see that LIME loses Robustness the more features we use due to the fact that we have more granularity over features.

\subsection{Dependence of LIME on the number of black box evaluations}
\label{analisisblackbox}

% En esta sección evaluaremos cuanto importa el número de evaluaciones de la caja negra. Este estudio es muy importante ya las evaluaciones de la caja negra son el cuello de botella más grande de la explicabilidad de caja negra. Aunque lo deseable es poder evaluar la función objetivo el máximo de veces posible, debe existir un balance entre la calidad de la explicación y el tiempo que se tarda en generarla.

In this section we will evaluate how important the number of black-box evaluations is over the LIME methods. This study is critical since black-box evaluations are the biggest bottleneck of black-box explainability methods. Although it is desirable to be able to evaluate the black-box function as many times as possible, there must be a trade-off between the quality of the explanation and the time it takes to generate it.

\paragraph{\textbf{Local Concordance}}

In Figure \ref{fig:concordance_ME}, we can appreciate that increasing the number of black-box evaluations does not change the Local Concordance score significantly. Also, if we look at absolute values, we realize that we obtain significantly high values. This is due to the fact that the sampling used by LIME is very stable in picking the neighbors close to the original example. 

% Podemos apreciar que el aumento de número de evaluaciones de la caja negra no hace variar demasiado la puntuación en Local Concordance. Esto se debe a que el muestreo utilizado por LIME es muy estable escogiendo los vecinos cercanos al ejemplo original. 

\begin{figure}[t]
    \centering
    \includegraphics[height=0.5\textwidth]{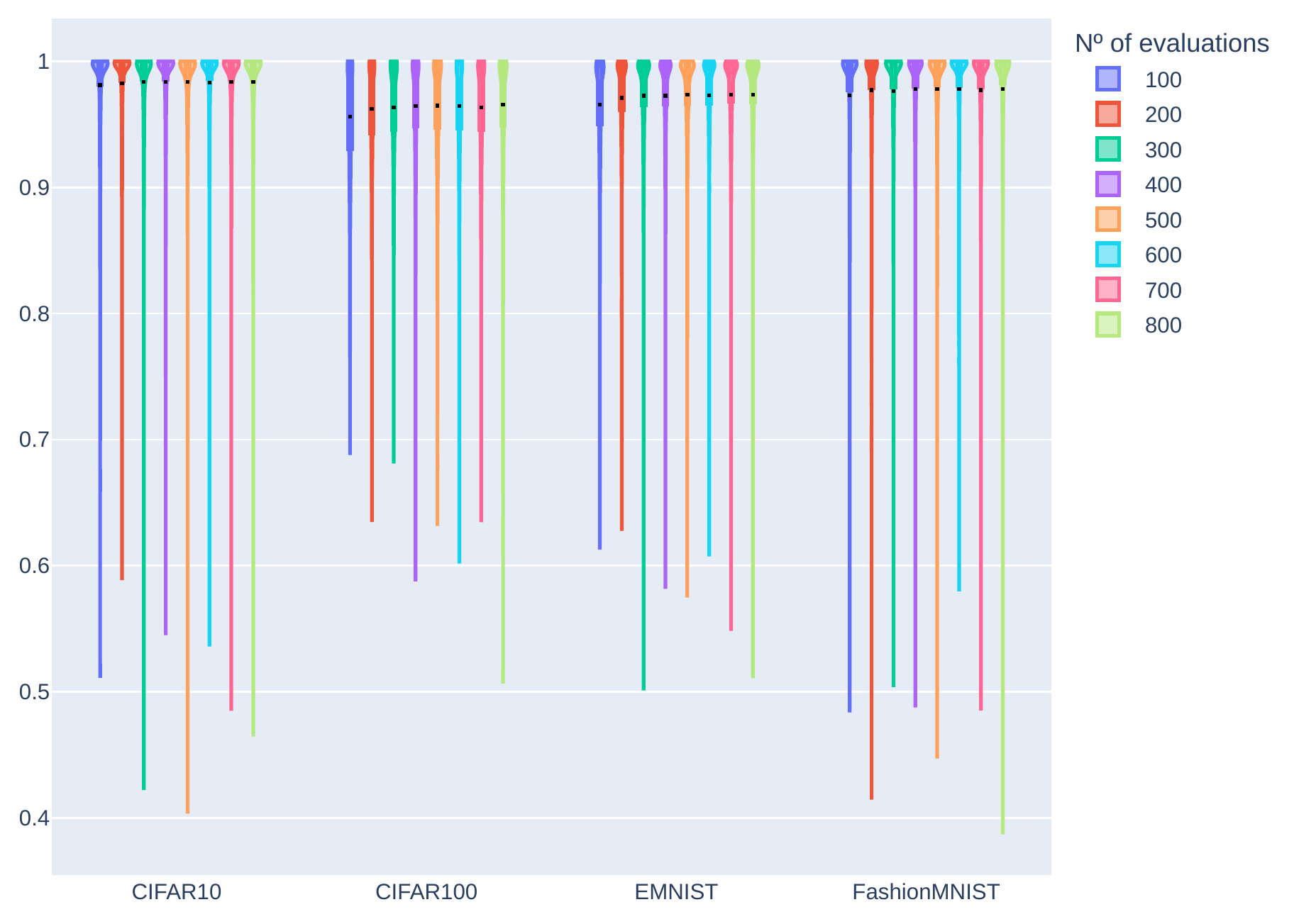}
    \caption{Performance of LIME methods grouped over max number of evaluations in the \textbf{Local Fidelity} metric.}
    \label{fig:fidelity_ME}
\end{figure}

\paragraph{\textbf{Local Fidelity}}

In Figure \ref{fig:fidelity_ME}, we appreciate that, in this case, the more evaluations of the black-box, the better result. We obtain marginally the neighbors close to the original example appear less frequently than the original example. We may expect that by randomly generating more neighbors we obtain a better score in the neighborhood of the original example.

% Apreciamos que, en este caso, cuantas más evaluaciones de la caja negra obtenemos un resultado ligeramente mejor. Esto se debe a que los vecinos cercanos al ejemplo original aparecen de manera menos frecuente que el ejemplo original. Es natural que al generar aleatoriamente más vecinos obtengamos mejor puntuación en el vecindario del ejemplo original.

\begin{figure}[t]
    \centering
    \includegraphics[height=0.5\textwidth]{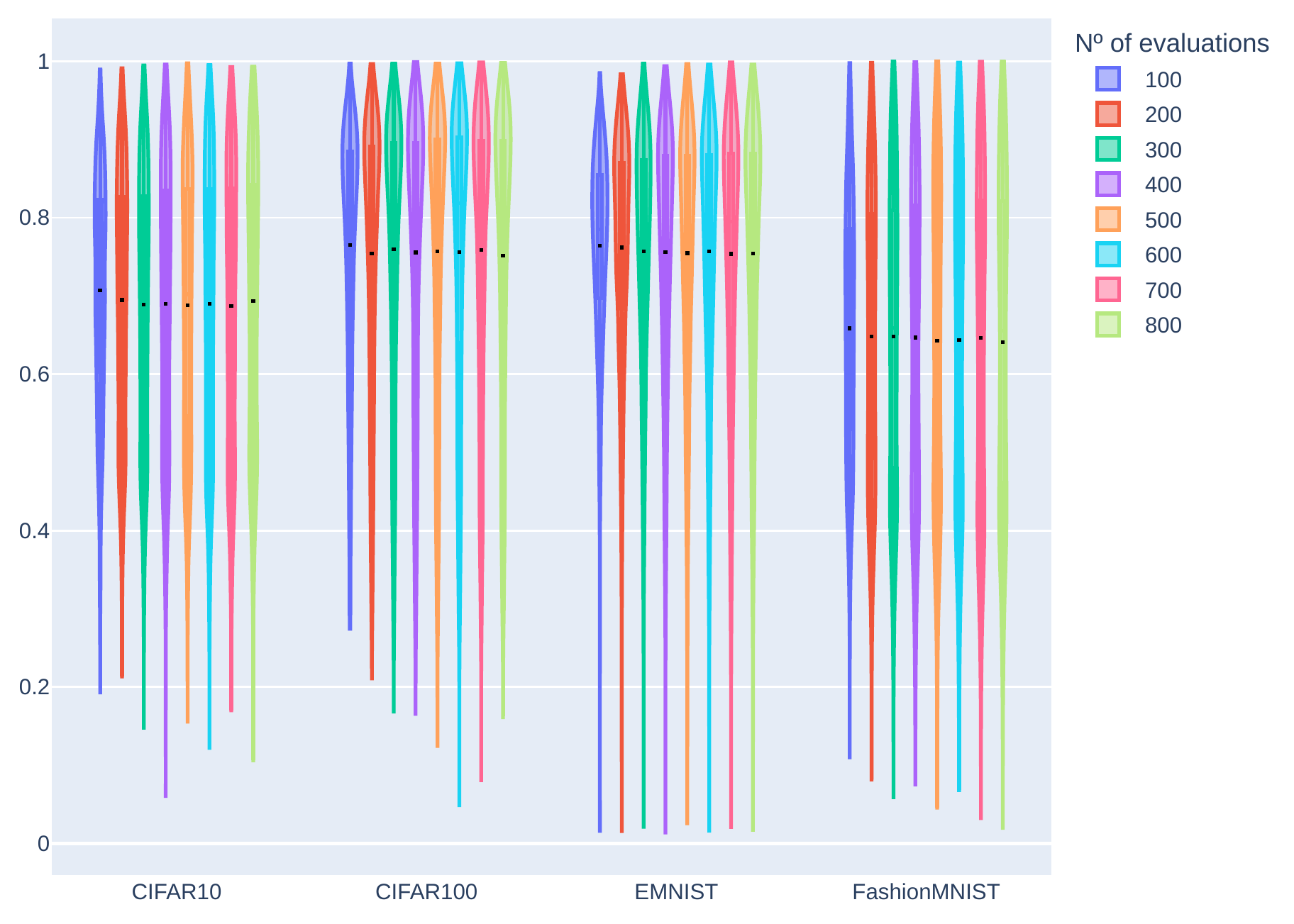}
    \caption{Performance of LIME methods grouped over max number of evaluations in the \textbf{Prescriptivity} metric.}
    \label{fig:prescriptivity_ME}
\end{figure}

\begin{figure}[t]
    \centering
    \includegraphics[height=0.5\textwidth]{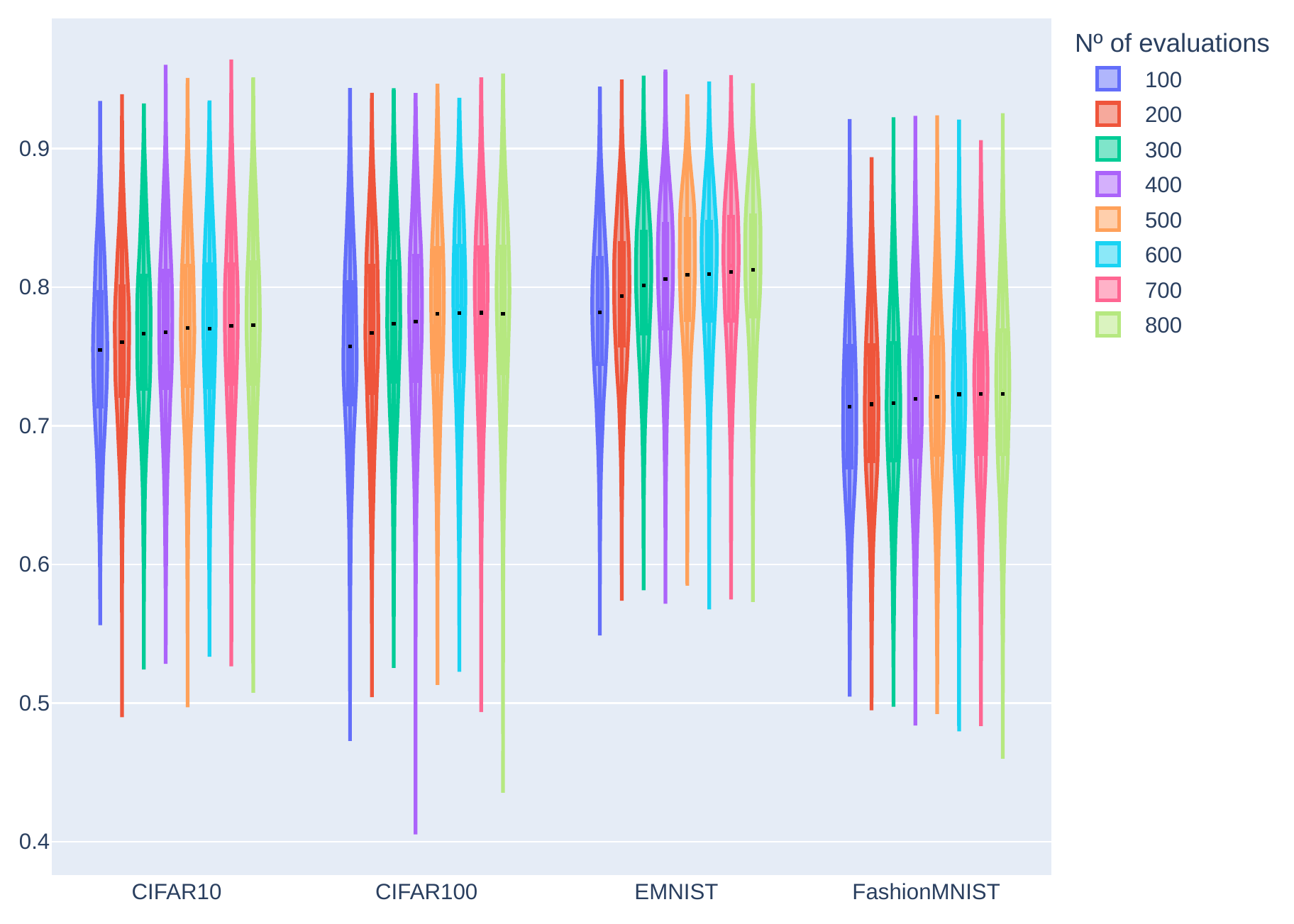}
    \caption{Performance of LIME methods grouped over max number of evaluations in the \textbf{Conciseness} metric.}
    \label{fig:conciseness_ME}
\end{figure}

\begin{figure}[t]
    \centering
    \includegraphics[height=0.5\textwidth]{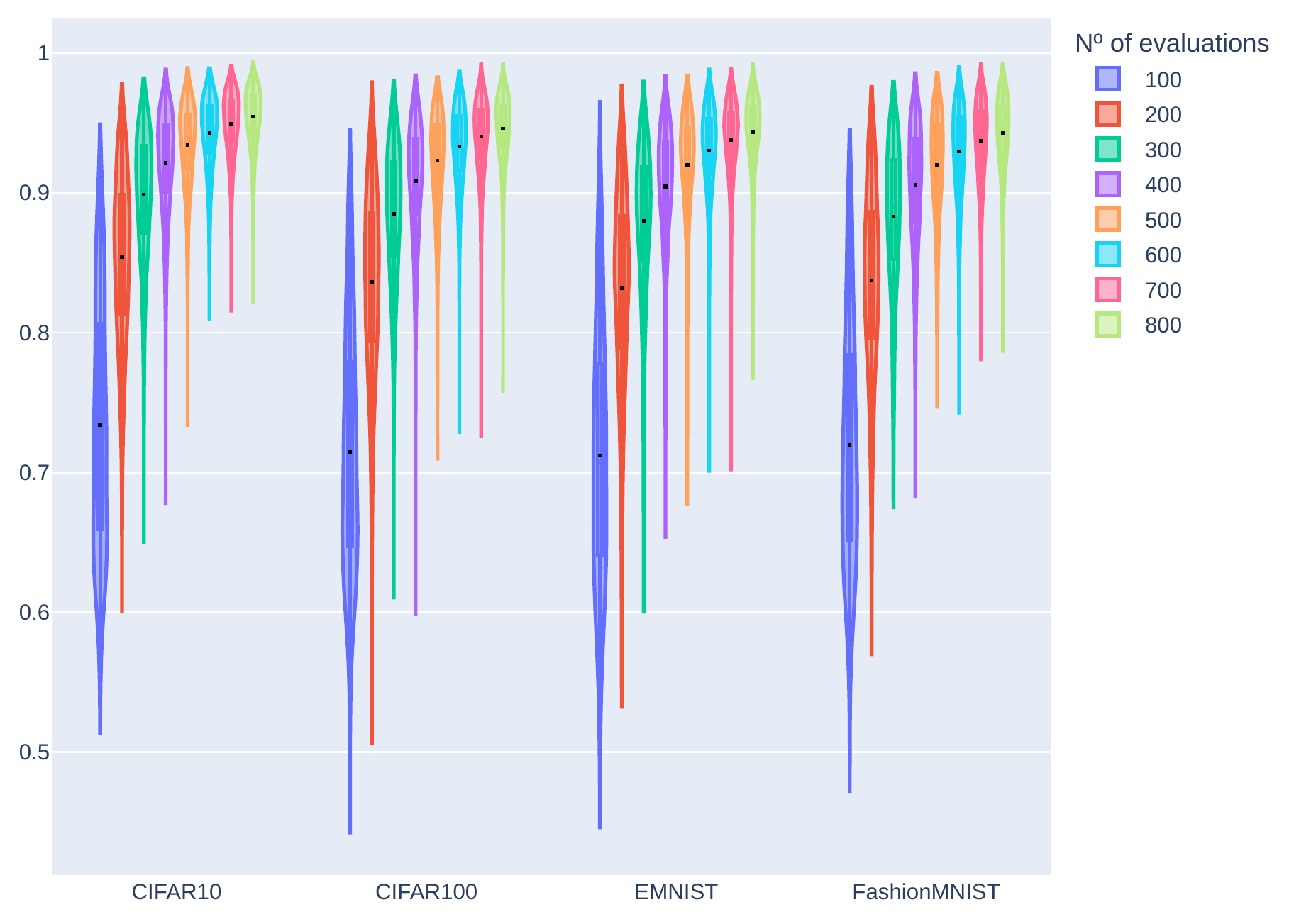}
    \caption{Performance of LIME methods grouped over max number of evaluations in the \textbf{Robustness} metric.}
    \label{fig:reiteration_ME}
\end{figure}

\paragraph{\textbf{Prescriptivity}}

In Figure \ref{fig:prescriptivity_ME}, we observe that the number of evaluations is not a differentiating factor. LIME proposes a series of changes that consistently change the prediction of the model by the same amount approximately.

% Vemos que no hay demasiadas diferencias entre utilizar más o menos evaluaciones, es decir, LIME propone una serie de cambios que, de manera consistente, cambian la predicción del modelo.

\paragraph{\textbf{Conciseness}}

In Figure \ref{fig:conciseness_ME}, we observe that the Conciseness metric is influenced by the number of evaluations of the black-box, making it less variable. Thus, LIME methods propose on average the same percentage of important features although increasing the number of evaluations tends to obtain less variable results, which is the main goal of increasing the number of maximum evaluation of black-box evaluations.

% Apreciamos que la métrica de concisión se ve influenciada por el numero de evaluaciones de la caja negra haciendola menos variable. Esto quiere decir que los métodos LIME proponen en media el mismo porcentaje de características importantes aunque al aumentar el número de evaluaciones tiende a obtener resultados menos variables. 

\paragraph{\textbf{Robustness}}

In Figure \ref{fig:reiteration_ME}, we observe that as the number of black-box evaluations increases, LIME methods become more consistent, although at the cost of using more computational time. Depending on the desired Robustness or time limit requirements, we can estimate of how much an explanation can change. 

\paragraph{\textbf{Global conclusion}}

In this case, the metric of Robustness is the one that outstands the most. Such results are expected since the more examples we use from the neighborhood, the less variable the generated explanation will be. Thanks to this analysis, we will be able to see what is the cost in time associated with a particular Robustness.

\subsection{LIME vs SHAP: General analysis over the explanation generators}
\label{analisisLIMEvsSHAP}

On this subsection we evaluate the performance on each proposed metric of LLE methods, LIME with $\sigma=2,3,4,5,6,7,8$ and SHAP, local and global versions. For this comparison, we considered the results of the above scenarios to choose the best number of features and the maximum number of black-box evaluations considered. In our case, we pick 64 features and 800 black-box evaluations.

\begin{figure}[t]
    \centering
    \includegraphics[height=0.5\textwidth]{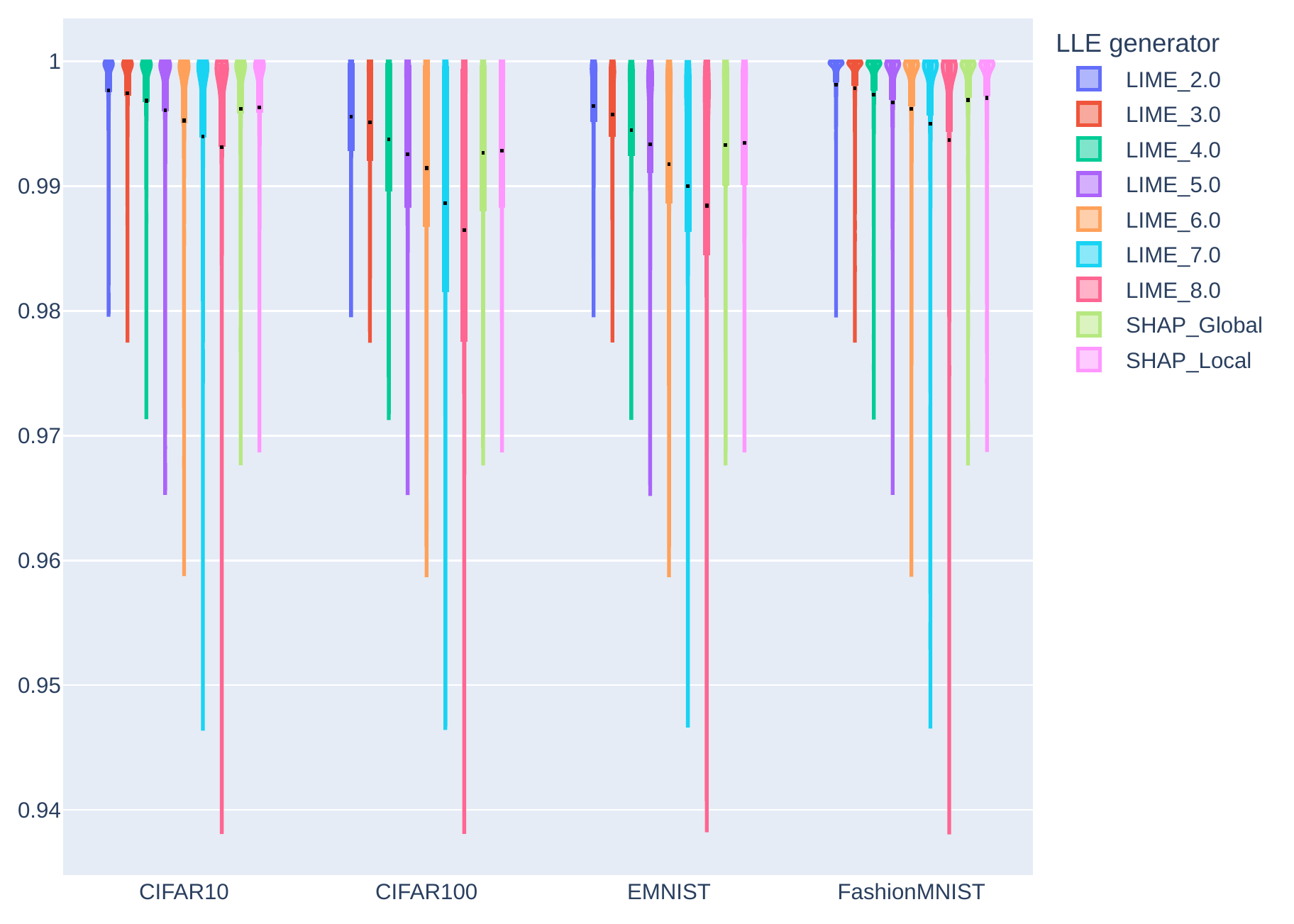}
    \caption{Performance of each explanation generator over the \textbf{Local Concordance} metric}
    \label{fig:concordance}
\end{figure}

\begin{figure}[t]
    \centering
    \includegraphics[height=0.5\textwidth]{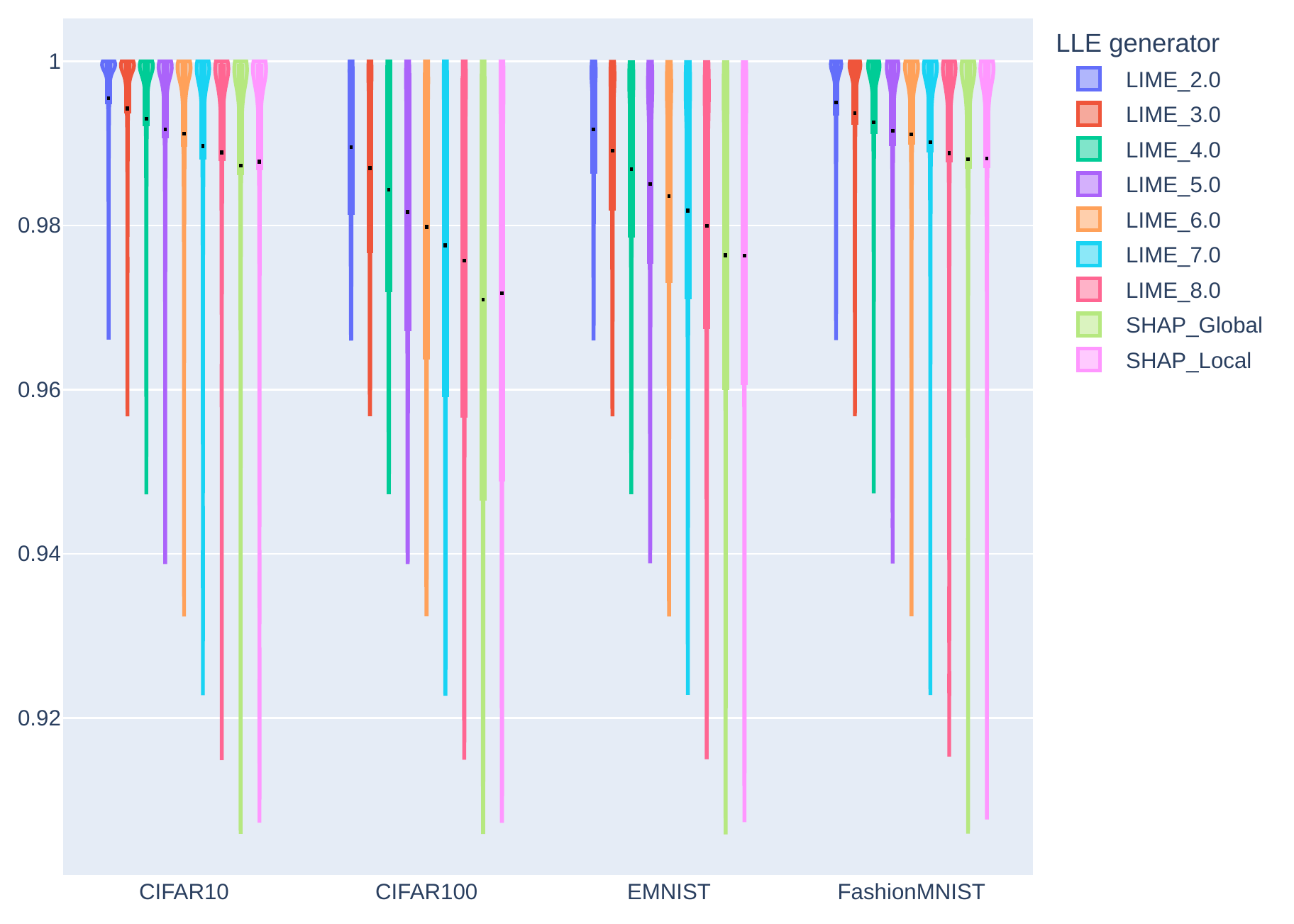}
    \caption{Performance of each explanation generator over the \textbf{Local Fidelity} metric}
    \label{fig:fidelity}
\end{figure}

\begin{figure}[t]
    \centering
    
    \includegraphics[height=0.5\textwidth]{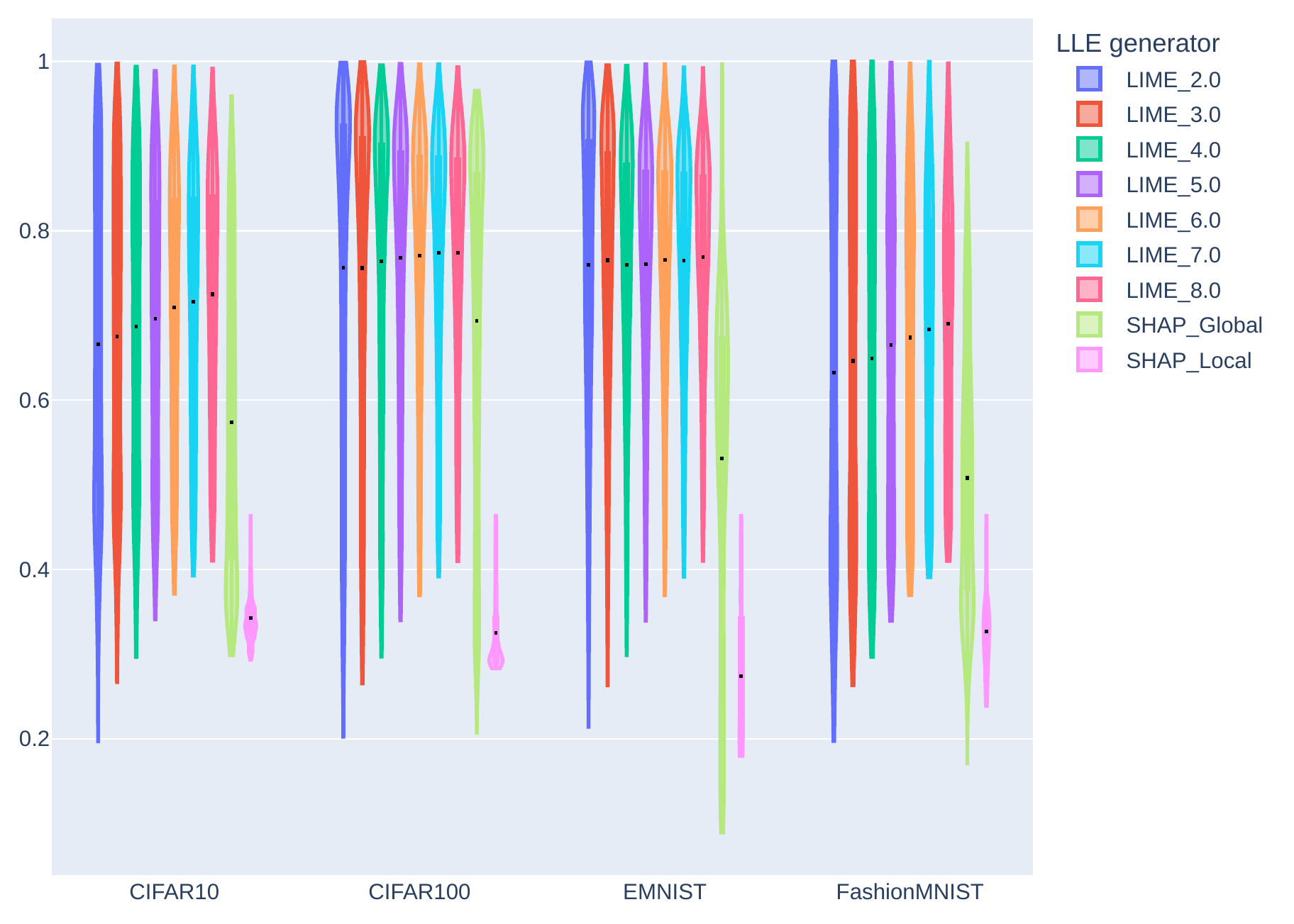}
    
    \caption{Performance of each explanation generator over the \textbf{Prescriptivity} metric}
    \label{fig:prescriptivity}
    
\end{figure}

\paragraph{\textbf{Local Concordance}} In Figure \ref{fig:concordance} we show the performance of the Local Concordance metric over all dataset. We observe that LIME with larger $\sigma$ perform worse. $\sigma$ parameter controls the width of the neighborhood generated, making the original example $x$ less relevant. On the other hand, local and global SHAP obtain stable and comparable results to those obtained by LIME with $\sigma=2,3,4$ because in each SHAP regression the relative importance of the original example $x$ remains constant with respect to the rest of the generated neighbors.

\paragraph{\textbf{Local Fidelity}} On Figure \ref{fig:fidelity} we note the same behavior for LIME methods as for the Local Concordance metric, i.e., the score of this metric decreases as $\sigma$ is higher since the larger the neighborhood it generates, the less importance is given to the direct surroundings of the $x$ example. We also note that SHAP methods obtains a worse result than LIME with $\sigma= 4$. This would mean that the behavior of SHAP gets worse as it moves away from the original $x$ example.

\paragraph{\textbf{Prescriptivity}} In Figure \ref{fig:prescriptivity}, we note that different LIME methods show similar performance regardless of $\sigma$, with slight variations between datasets. On the other hand, there is a noticeable loss in SHAP Local. This is partly due to the fact that SHAP gives significant weight to the original example $x$ when there is a large number of features and does not extrapolate to more distant examples. On the other hand, global SHAP performs slightly worse than LIME methods. It pays attention not only to the closest examples to the original $x$ example, but also to the farthest possible examples.

\begin{figure}[t]
    \centering
    \includegraphics[height=0.5\textwidth]{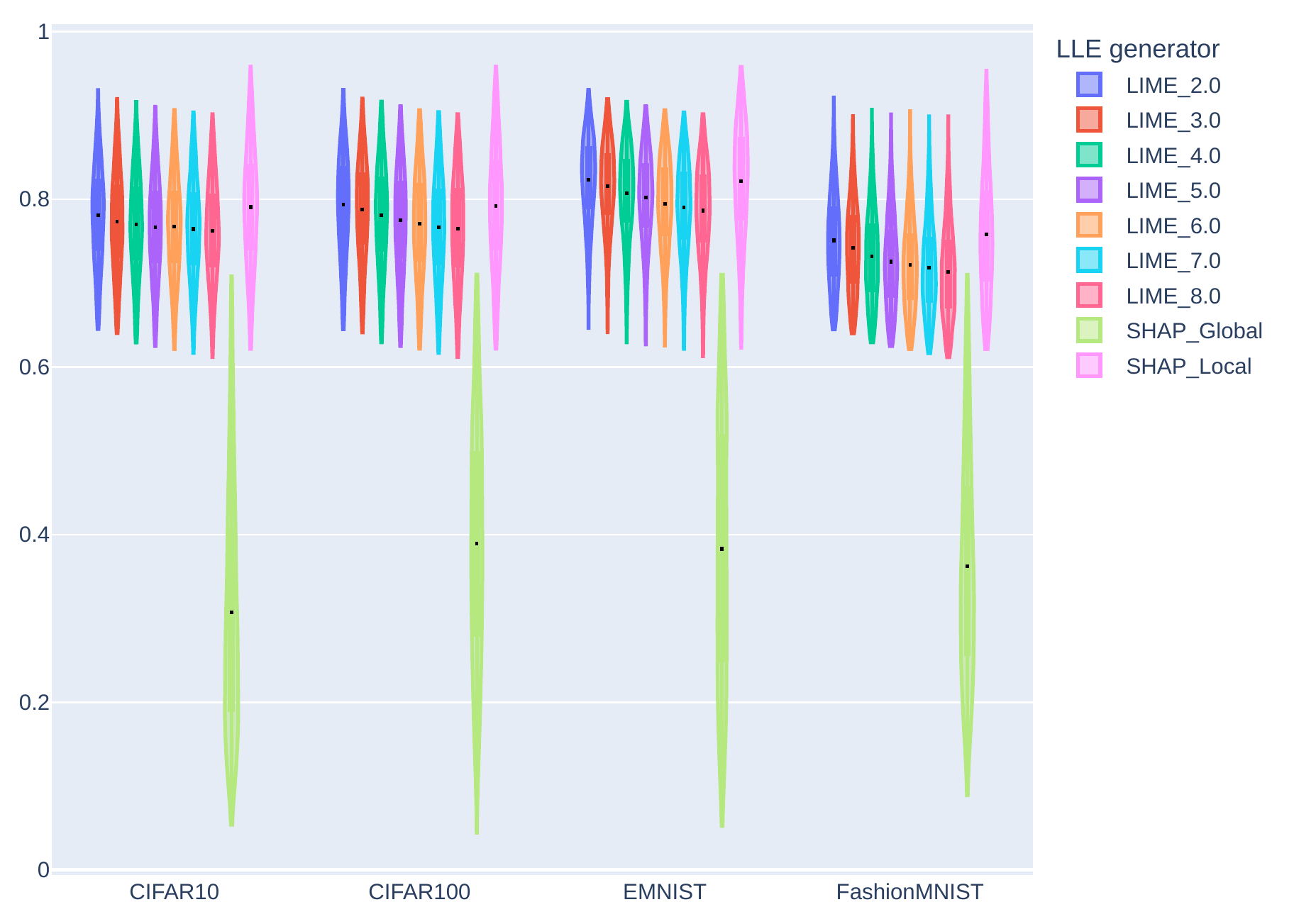}
    \caption{Performance of each explanation generator over the \textbf{Conciseness} metric}
    \label{fig:conciseness}
\end{figure}

\begin{figure}[t]
    \centering
    \includegraphics[height=0.5\textwidth]{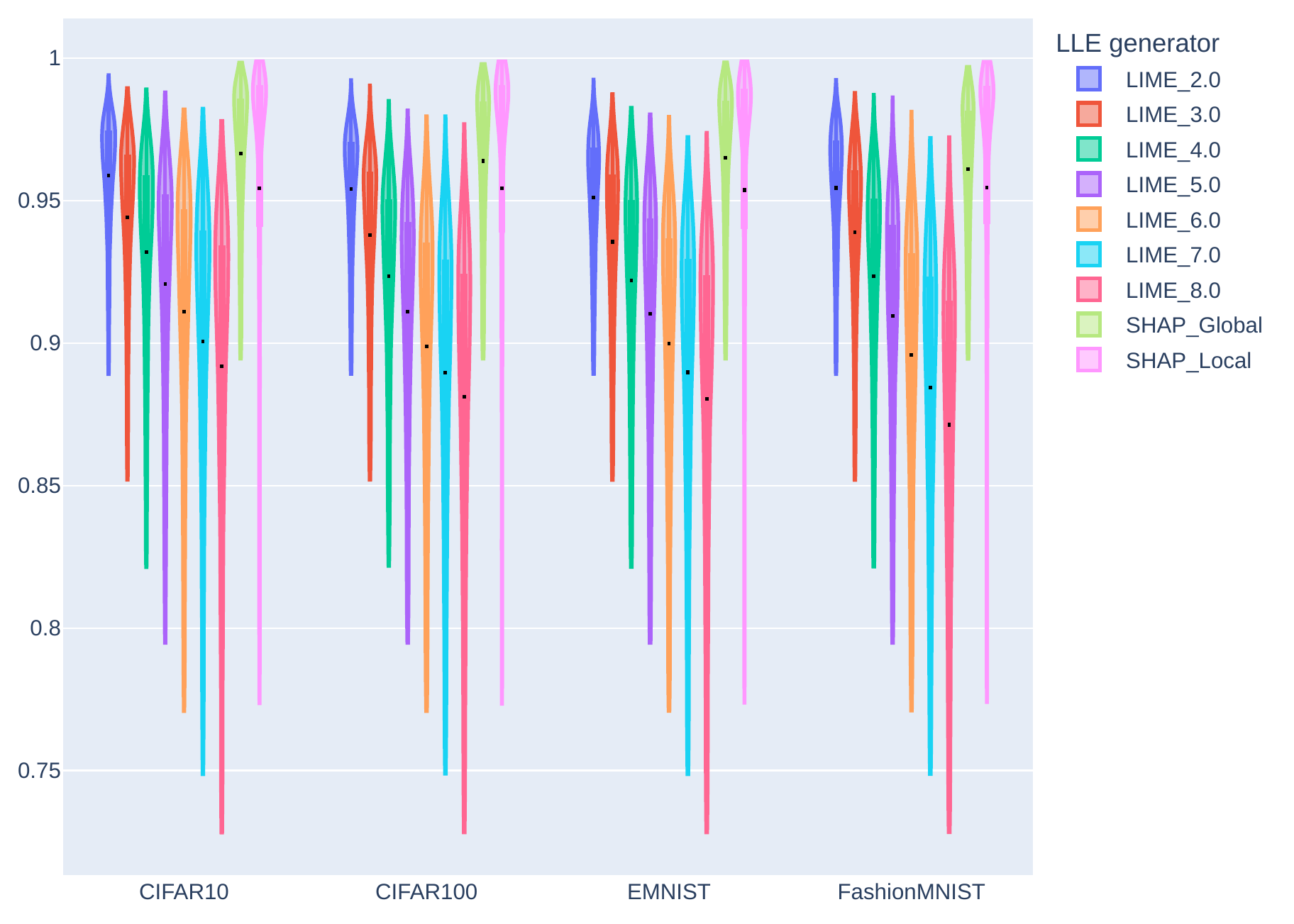}
    \caption{Performance of each explanation generator over the \textbf{Robustness} metric}
    \label{fig:reiteration}
\end{figure}

\paragraph{\textbf{Conciseness}} In Figure \ref{fig:conciseness} we note that the LIME methods have a similar behavior among the different $\sigma$ configurations, obtaining slightly different results depending on the dataset. On the other hand, the global SHAP method shows worse results, which tells us that SHAP global spreads its attention over too many features. On the other hand, local SHAP obtains a comparable score with the different LIMEs, which means that both methods spread its attention over almost the same number of features.

\paragraph{\textbf{Robustness}} In Figure \ref{fig:reiteration} we note that the best scoring results are obtained in this case by the SHAP models. This is due to the fact that SHAP methods choose neighbors in a stable way. LIME methods generate examples less stably as we increase the $\sigma$ parameter. The reason of the increase of $\sigma$ is that we also increase the size of the neighborhood and, therefore, the diversity of the generated neighbors.

\subsection{Global analysis and lessons learned}

% Una vez analizados los comportamientos de cada métrica por separado, podemos extraer lecciones aprendidas sobre cada uno de los métodos evaluados gracias al potencial para auditar del framework REVEL. 

Once we have analyzed the performance of each metric separately, we can extract lessons learned about each of the methods evaluated thanks to the auditing potential of the REVEL framework.

\begin{itemize}
    \item \textbf{SHAP:} It focuses too much on the concrete example to be explained and does not generalize well in the synthetic neighborhood. Local Concordance is good although the Local Fidelity, in comparison with LIME, is worse than expected and Prescriptivity results are very poor. Although they are very stable methods, as we observe in the Robustness metric, we may establish, in conjunction with the previous conclusions, that they are in fact methods whose neighborhood is too small and therefore they use almost all the same examples to generate explanations. 
    
    \item \textbf{Global SHAP vs Local SHAP:} The main difference between Local and Global SHAP is found in Prescriptivity and Conciseness. Local SHAP is able to discard unimportant features, while Global SHAP hardly does so. The reason for this behavior is because Local SHAP is using only the neighborhood near the instance to be analyzed, while Global SHAP  uses also the instances of completely empty images except for some particular patch. In other types of data, this approach is correct (e.g., in tabular data, to see if any particular feature biases the overall result) but in the case of images, an almost entirely gray image does not give much information.

    \item \textbf{LIME:} This method focuses on the local neighborhood of the example to be explained. We observe that the parameter $\sigma$ establishes the size of the neighborhood and, as it increases, it obtains worse results in the local environment but has greater generalization power. We deduce this because in the metrics of Local Concordance and Local Fidelity it worsens with increasing $\sigma$ but remains stable or even increases in Prescriptivity. The increase in neighborhood size also results in slightly more attention being paid to diverse features and, in addition, causes a more diverse generation of neighbors, as we see in the Conciseness and Robustness metrics respectively.

\end{itemize}

In conclusion, we may establish that SHAP focuses too much on the example to be explained while LIME is able to generalize better on these datasets.

% Por último, la lección aprendida más importante es el estudio exhaustivo y matemáticamente robusto que se ha realizado para el desarrollo de REVEL. Gracias al mismo, no solo hemos podido establecer medidas comparativa entre explicaciones, sino que además estas medidas sirven de manera absoluta, sin necesidad de comparar con otras explicaciones. 

Finally, the most important lesson learned is the exhaustive and mathematically robust study we performed for the development of REVEL. Thanks to this study, we have not only been able to establish comparative measures between explanations, but also that these measures serve as absolute measures, without the need to compare with others.

%% file: sections/6_Conclusion.tex
\section{Concluding remarks}
\label{conclusion}

In this paper we present REVEL, a novel framework specialized in analysis and comparison of explanations. We provide a theoretical guideline for the use of REVEL. We also provide a practical illustration of usage of REVEL by comparing LIME and SHAP methods in four different benchmarks. 

As lessons learned over, we want to remark that having bounded metrics with well-defined limits gives us absolute information on every evaluation aspect and not only a comparative one. This is useful to dismiss explanations by themselves even if there is no baseline to compare with. For the development of future metrics, this characteristic is desirable. 

Regarding the developed metrics themselves, we can extract the following lessons: Local metrics can help us to detect biases comparing with Prescriptivity, Conciseness provides us information about whether an explanation is useful or not by the percentage of discarded features, Robustness shows information on the stability of the explanations.

From the above analysis, we can establish that, within the black-box methods of explanation proposals over the image classification task, LIME behaves better than SHAP because SHAP focuses too much on the locality of the example to be explained, while LIME is able to generalize much better.

Once the method of explanation has been chosen for a particular model, we emphasize that the analysis should not stop there but analyze different aspects such as the number of features considered or the number of evaluations of the black-box necessary for a robust explanation.